\documentclass[12pt, final]{l4dc2020} 

\usepackage[frozencache]{minted}
\definecolor{mintedbackground}{rgb}{0.98,0.98,0.98}
\setminted{framesep=2mm,bgcolor=mintedbackground, fontsize=\scriptsize,linenos} 

\usepackage{etoolbox}
\BeforeBeginEnvironment{wrapfigure}{\setlength{\intextsep}{0pt}}

\usepackage{xspace}
\newcommand{\method}[1][]{plan2vec\xspace}
\newcommand{\Method}[1][]{Plan2vec\xspace}

\usepackage{times}
\usepackage[utf8]{inputenc} 
\usepackage[T1]{fontenc}    
\setcitestyle{numbers,square}
\usepackage{nameref}
\usepackage{hyperref}       
\AtBeginDocument{\let\latexlabel\label}

\usepackage{mathtools}

\usepackage{booktabs}       
\usepackage{nicefrac}       
\usepackage{microtype}      
\usepackage{algorithm}
\usepackage[noend]{algorithmic}
\usepackage{caption}
\usepackage{subcaption}
\usepackage{mathrsfs}
\usepackage{enumitem}
\usepackage{xcolor}
\usepackage{wrapfig}
\usepackage{listings}
\usepackage{sidecap}
\usepackage{makecell}

\newcommand{\comment}[1]{\iffalse #1 \fi}

\newcommand{\fig}[1]{Fig.~\ref{#1}}

\newcommand{\tab}[1]{Table~\ref{#1}}
\newcommand{\alg}[1]{Algorithm~\ref{#1}}

\definecolor{es-green}{HTML}{3dc466}
\definecolor{es-blue}{HTML}{23AAFF}
\definecolor{es-red}{HTML}{EF8582}
\definecolor{es-gray}{HTML}{757575}

\usepackage{amsmath,amsfonts,bm}

\def\eqref#1{equation~\ref{#1}}

\def\1{\bm{1}}

\DeclareMathAlphabet{\mathsfit}{\encodingdefault}{\sfdefault}{m}{sl}
\SetMathAlphabet{\mathsfit}{bold}{\encodingdefault}{\sfdefault}{bx}{n}

\DeclareMathOperator*{\argmin}{arg\,min}

\title{Plan2vec: Unsupervised Representation Learning by Latent Plans}

\author{%
    \Name{Ge Yang}\(^{*\dagger}\) \Email{ge.ike.yang@gmail.com}\\
    \Name{Amy Zhang}\(^{*\dagger}\) \Email{amyzhang@fb.com}\\
    \Name{Ari S. Morcos}\(^\dagger\) \Email{arimorcos@fb.com}\\
    \Name{Joelle Pineau}\(^{\dagger\ddagger}\) \Email{jpineau@cs.mcgill.ca}\\
    \Name{Pieter Abbeel}\(^\mathsection\) \Email{pabbeel@cs.berkeley.edu}\\
    \Name{Roberto Calandra}\(^\dagger\) \Email{rcalandra@fb.com}\\
    \addr \(^\dagger\)Facebook AI Research, \(^\ddagger\)McGill University, \(^\mathsection\)UC Berkeley\\%
}

\begin{document}

\maketitle

\begin{abstract}
In this paper we introduce \method, an unsupervised representation learning approach that is inspired by reinforcement learning. \Method constructs a weighted graph on an image dataset 
using near-neighbor distances, and then extrapolates this local metric to a global embedding by distilling path-integral over planned path.
When applied to control, \method offers a way to learn goal-conditioned value estimates that are accurate over long horizons that is both compute and sample efficient. We demonstrate the effectiveness of \method on one simulated and two challenging real-world image datasets. Experimental results show that \method successfully amortizes the planning cost, enabling reactive planning that is linear in memory and computation complexity rather than exhaustive over the entire state space. Additional results and videos can be found at \url{https://geyang.github.io/plan2vec}.

\end{abstract}

\section{Introduction}

A good representation of the state space is essential to an intelligent agent that is trying to accomplish tasks in the world. For this reason, we look at representation learning through the lens of reinforcement learning. Under the standard Markov decision process (MDP,~\citealt{Bellman1957}) formulation, the state space \(S\) appears as the input domain for 
two types of functions. The first type is \textit{local}, such as the transition probability \(\mathbb P\) and the step-wise reward \(R\). The second type is \textit{non-local} and requires integration along paths, such as the state value \(V(s)\) or the \(Q\)-function~\citep{haarnoja2018soft,silver2014deterministic,lillicrap2015continuous,abdolmaleki2018maximum}. For a specific type of task that can be formulated as accomplishing goals~\cite{kaelbling1993learning}, the goal-conditioned value function \(V(s, g)\) becomes a (negative) metric. The very focus of modern reinforcement learning is to learn \(V\), for it parametrically encodes optimal plans. In policy search, such distance function acts are useful as a shaped reward~\cite{haarnoja2018soft,hartikainen2019dynamical,upn}.

In this paper, we ask the question: is there a way to learn this type of metric representation without explicitly involving interactions with the environment, using only offline exploratory data that are abundantly available? Our key insight is that one can remove the need for learning dynamics by modeling these \textit{local} relationships between near-neighbors as the edges of a graph, then use heuristic search to generate optimal long horizon plans for path integration. Our proposed method -- \textit{\method}\,-- appears in two variants: the first uses regression towards a planned trajectory similar to dynamic distance learning \cite{hartikainen2019dynamical} but with a strong search expert, whereas the second uses fitted value iteration \citep{gordon1995stable,ernst2005tree,riedmiller2005neural}.

To help illustrate our method, we lead the introduction of \method with a set of simulated visual navigation tasks. We show the importance of using graph search as a sampling policy as opposed to a memory-less planner by looking at the planning success rate for k-steps of plan-ahead~(see~\fig{fig:planning_and_sample_complexity}). Then we demonstrate our approach on deformable object manipulation with a rope dataset~\citep{wang2018ropedataset} that is otherwise difficult to model. Finally, we tackle a challenging real-world navigation dataset Street Learn~\citep{streetlearn} to show that \method is able to learn to navigate from sequences of Street View images driving through the streets, with \textit{no} access to the ground-truth GPS location data. Interestingly, we found that the representation \method learns contains an interpretable metric map, despite that the graph it distills from is topological in nature.

\begin{figure}[t]
    \centering
    \includegraphics[width=1\textwidth]{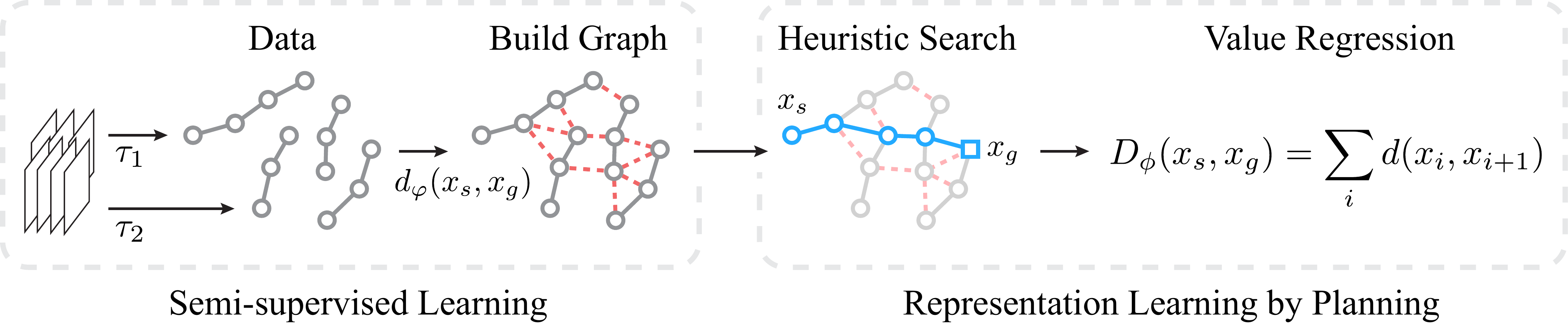}
    \caption[Plan2Vec]{High-level schematics of plan2vec. The dataset contains sequences of observations. The graph building step can be considered semi-supervised learning, where the real transitions are the labeled data, and the task is to find new transitions by learning a local metric \(d\) that generalizes. The representation learning step can be considered learning an embedding \(\phi\) of the graph. Plan2vec uses plans made on the graph to generate value targets for \(D\), the \textit{shortest-path-distance} metric.}
    \label{fig:plan2vec_main}
\end{figure}

\section{Technical Background}
The goal of reinforcement learning is to find a policy distribution \(\pi(a \vert s)\) that maps from the state space \(S\) to the action space \(A\) for a given Markov decision process (MDP)~\citep{Bellman1957}, such that it maximizes the return, defined as the discounted sum of future rewards \(J_\pi = \sum_t \gamma^t R(s_t, a_t)\). The optimality of a policy is provided by the Bellman equation%
\begin{equation}
V(s) = \mathcal T V^*\textnormal{,\quad where\quad} \mathcal T V^* \equiv R(s, a, s') + \max_a \sum_{s'} \mathbb P(s'\vert s, a) \;\gamma V^*(s').\label{eq:bellman}
\end{equation}%
\(\mathbb P(s'\vert s, a)\) is the transition probability. \(\mathcal T\) is the contraction operator defined recursively on the state-value function \(V(s)\). We further assume that the MDP is fully observable, so there \(\exists\) a mapping \(\phi(o_s) \mapsto z_s\) from the space of observations \(O\) to a latent space \(Z\), for each state \(s\).

When deep neural network is used as a function approximator~\cite{mnih2013dqn}, learning is typically implemented as sample-based regression towards an n-step bootstrapped value target~\citep{mnih2016asynchronous}
\begin{equation}
\mathcal L = \left\Vert V(s_0) - \sum_{t=0}^n \gamma^{t-1} r - V^*(s_n)\right\Vert_2.
\end{equation}

\paragraph{Generalized Value Function as A Metric} 
Learning to achieve goals is an important subproblem of reinforcement learning~\citep{kaelbling1993learning}. In a goal reaching task, the agent incurs a \textit{cost} of \(- d(s, s')\) at each step. The \textit{distance-to-goal} \(D(s, g)\) refers to the \textit{shortest path distance} \(\min_\tau \sum_{x\sim \tau} d(x_i, x_{i + 1}) \) between \(s\) and \(g\). This formulation offers additional structure in that \(D\) is a metric that satisfies the triangular inequality
\begin{equation}
\forall s' \in \mathcal S,\;  D(s, g) \leq D(s, s') + D(s', g)\,.
\end{equation}%
For this reason, the \textit{generalized value function}~(GVF,~\citealt{sutton2011horde}) family of algorithms~\citep{littman2001predictive,schaul2015universal,kulkarni2016successor,singh2003predictive} formulate learning a goal-conditioned Q value function as learning \textit{predictive features}. For our purpose of doing unsupervised representation learning without actions, this can be simplified as learning a value  \(V(o, o_g) = - D_\phi(o, o_g)\) where \(D_\phi(o, o_g) \equiv \Vert \phi(o),\; \phi(o_g) \Vert_p\) is the distance between the latent features vectors.

\paragraph{Dataset as A Graph}
For a dataset of images \(\{x_i\}\) there \(\exists\) a weighted graph \(\mathcal G = \left\langle\mathcal V, \mathcal E\right\rangle\) where each vertex \(v_i\) corresponds to an image \(x_i\). \(e_{ij} \in \mathcal E \) iff according to a local metric \(d\), \(d(x_i, x_j) < d_0\). We let the edge \(e_{ij}\) weight by \(w_{ij} = d(x_i, x_j)\). If we make the additional assumption that the data are sequences of observations, then the graph is directed.

\section{Unsupervised Representation Learning by Latent Planning}
\begin{wrapfigure}{r}{.27\textwidth}
    \centering
    \includegraphics[width=0.95\linewidth]{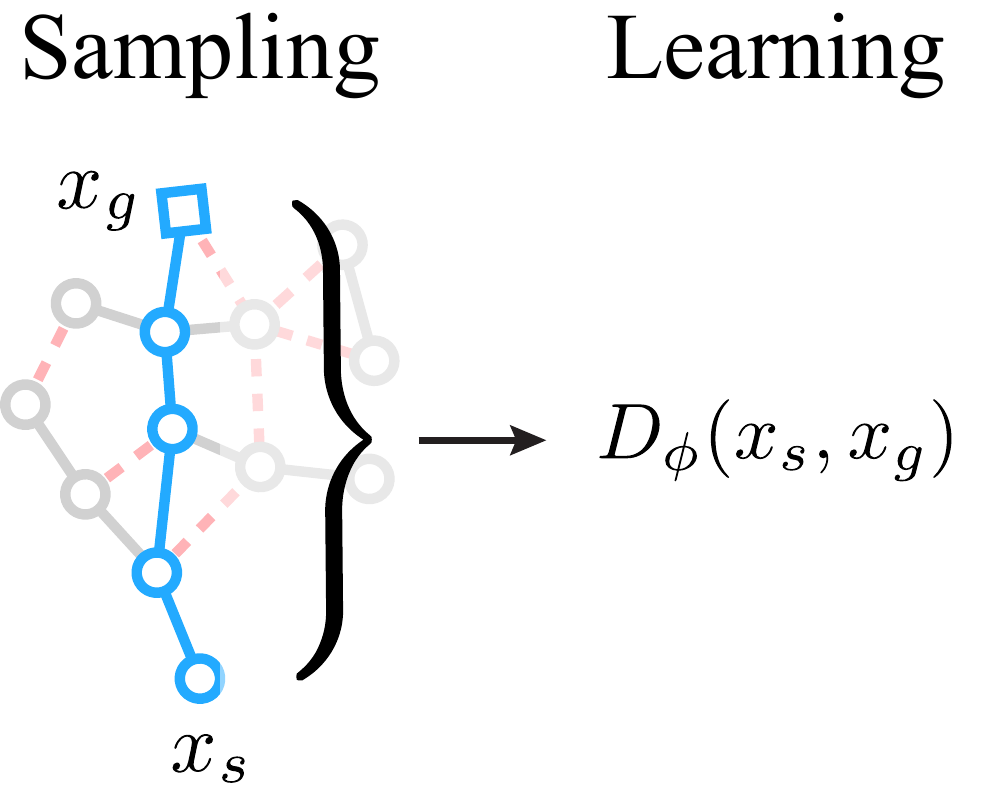}
    \caption{Plan2vec uses planning to generate value targets for the metric \(D_\phi\)}\label{fig:sample_and_learning}%
\end{wrapfigure}
Plan2vec is built upon the idea that for a collection of images with a local metric \(d\), the graph \(\mathcal G\) weighted by \(d\) is embedded by a Riemann manifold, the metric of which is the shortest-path-distance \(D\). By choosing the function class \(D_\phi\) that decomposes into an embedding function \(\phi(x)\) and a metric \(\Vert \cdot \Vert_p\), we project \(D\) to a \(\ell^p\)-metric space with a vector embedding. 

\paragraph{Problem Formulation}

Plan2vec models the representation learning problem as learning how to play a goal-reaching-game in which one is tasked to find the shortest path from one observation to another, by hopping between near neighbors~(\fig{fig:sample_and_learning}). Under the context of reinforcement learning, \method treats the graph as a model of the state space, and learns from Dyna-styled unroll using graph-search as an expert policy~\citep{sutton1991dyna,anthony2017exit}. \Method treats the construction of the graph as a semi-supervised problem (\fig{fig:plan2vec_main}). It first adds transitions from the dataset as edges with weight \(1\). It then uses these as labeled data to learn a local metric \(d\), to generalize to other pairs of images as a form of loop closure. An edge \(e_{ij}\) is created between the node \(v_i\) and \(v_j\) if the distance between the corresponding images \(d(x_i, x_j) \leq d_0\), a hyper parameter.

\paragraph{Learning Local Metric}

Noise-contrastive estimation cast representation learning as maximizing the contrast between two distributions: the joint distribution between related views \(p(x, x^+)\), versus the product of the marginals \(p(x)p(x^-)\)~\citep{oord2018representation,gutmann2010noise,tian2019contrastive}
\begin{equation}
    L_\textnormal{NCE} = - \log \frac{\exp S(x, x^+)}{\exp S(x, x^+) + \sum_i^k \exp S(x, x_i^-)}\,,\label{eq:nce}
\end{equation}%
where \(S\) is the similarity function to be learned. \(\langle x, x^+ \rangle\) is the positive pair sampled in-context. \(\langle x, x_i^-\rangle\) is the negative pair sampled independently from the marginals. Under the context of control, Eq.\ref{eq:nce} has a natural interpretation as maximizing the log-probability that a pair \(\left\langle x, x^+\right\rangle\) is \textit{reachable} against pairs sampled at random, and is related to the \textit{distance metric} by \(d(x, x') \propto - \log p(x, x')\).

In the maze domain we directly regress the distance metric \(d\) towards one of~\{\textit{identical}, \textit{close}, or \textit{far-apart}\} with a nominal distance of \(\{0,\;1,\; 2\}\)
\begin{equation}
    \mathcal L_\textnormal{d} = \left\vert d(x, x) - 0 \right\vert + \left\vert d(x_t,\; x_{t+1}) - 1 \right\vert + \left\vert d(x,\; x^-) - 2 \right\vert.
\end{equation}%
When \(d_\phi\) is a Siamese network with an \(\ell^2\) metric, the first term can be dropped. We use smoothed \(L_1\) loss for all terms.

\begin{figure}[tb]%
\begin{minipage}[t]{0.42\textwidth}%
\begin{figure}[H]
    \centering
    \includegraphics[height=8.5em]{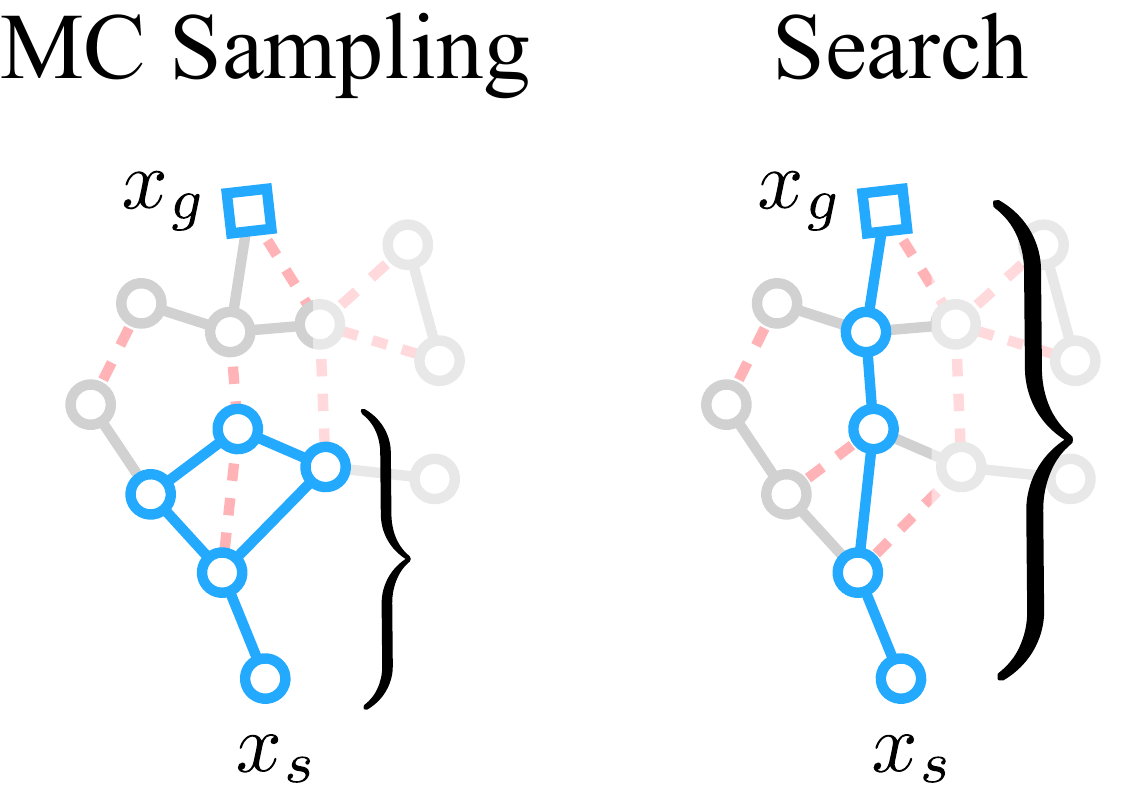}
    \caption{Memoryless sampling vs search.}
    \label{fig:sampling}
\end{figure}
\end{minipage}
\hfill
\begin{minipage}[t]{0.55\textwidth}%
\begin{algorithm}[H]
    \algsetup{linenosize=\footnotesize}
    \small
    \caption{\small \Method via Amortized Search}
    \label{alg:plan2vec_main}
    \begin{algorithmic}[1]
    \REQUIRE weighted directed graph \(\mathcal G = \langle \mathcal V, \mathcal E\rangle\) 
    \REQUIRE Shortest Path First search algorithm \(\mathrm{SPF}\)
    \STATE Initialize \(D_\Phi(x, x')\), let \(V = - D\).
    
    \WHILE{not converged}
        \STATE sample \(x_{v_s}, x_{v_g}\) and \(v_s, v_g\ \in \mathcal G\) as \textit{start} and \textit{goal}
        \STATE\label{step:alg_plan} find shortest plan \(\tau^* = \mathrm{SPF}(\mathcal G, v_s, v_g, D_\Phi)\)
        \STATE minimize \(
            \delta = \left\vert D_\Phi(x_s, x_g) - \sum_{v_i\sim \tau^*} d(x_{v_i}, x_{v_{i+1}}) \right\vert\)
    \ENDWHILE
    \end{algorithmic}
\end{algorithm}
\vspace{1em}
\end{minipage}
\end{figure}

\paragraph{Learning Representation by Latent Plans}\label{sec:amortize}
\Method samples pairs of images \(x_s\) and \(x_g\) and their corresponding vertices \(v_s\) and \(v_g\) from the graph \(\mathcal G\), then uses heuristic search to find the shortest path \(\tau^*\) in-between~(\alg{alg:plan2vec_main}).  Non-learning search algorithms typically discard the search tree after backtrack (step~\ref{step:alg_plan}). \Method collect these to generate regression targets for learning the value estimate with or without value bootstrapping. With the latter, one can use a fixed search depth \(h\).
\begin{equation*}%
\refstepcounter{equation}\latexlabel{eq:no_bootstrapping}%
\refstepcounter{equation}\latexlabel{eq:value_iteration}%
\underset{\strut\mathclap{\textnormal{no bootstrapping}}}{\mathrm{V(x_s, x_g)}} = - \sum_i d(x_{v_i}, x_{v_{i+1}}) 
\qquad
\underset{\strut\mathclap{\textnormal{bootstrapped}}}{\mathrm{V(x_s, x_g)}} = - \sum_{i=0}^{h-1} d(x_{v_i}, x_{v_{i+1}}) + V(x_{v_h}, x_{v_g}).
\tag{\ref*{eq:no_bootstrapping}, \ref*{eq:value_iteration}}
\end{equation*}
\noindent
We experimented with both fitted value-iteration (FVI) and amortized heuristic search for learning on a graph. The main short-coming with FVI is that relaxation for finding the shortest path occurs via gradient-based, iterative updates. Such scheme is unstable when applied to a graph as cycles within each rollout stall learning; whereas heuristic search explicitly avoid vertex-revisit at planning time.

\section{Related Works}

Plan2Vec builds upon two rich bodies of literature: unsupervised methods that learn an embedding from a local context, and value-based reinforcement learning methods that learn a policy.
In the first category, time-contrastive network (TCN), skip-gram (word2vec), contrastive predictive coding (CPC) and locally linear embeddings~\citep{sermanet2017time,mikolov2013efficient,oord2018representation,roweis2000nonlinear} are a family of methods that embed images, word tokens and image patches by making each sample similar to its neighbors in a small neighboring context. Similarly, graph embedding algorithms such as DeepWalk, Node2vec and diffusion maps~\citep{perozzi2014deepwalk,grover2016node2vec,socher2008manifold} randomly sample short trajectories in the neighborhood of a node to provide context. The locality of such context is restrictive, because one can not expect clear supervision from samples further apart. \Method solves this problem by replacing those random processes with graph-search to directly generate long-horizon distance targets between nodes that are arbitrarily far apart.

Embed to control (E2C), robust controllable embedding (RCE), L-SBMP and causal InfoGAN~\citep{watter2015embed,banijamali2017robust,ichter2018robot,kurutach2018learning} are a line of generative models that incorporate forward modeling in the latent space. They show that the learned representation is \textit{plannable}, but the models are limited to modeling local relationships. \Method differs by explicitly learning a \textit{shortest-path-distance} metric to embed the weighted graph on a Riemann manifold that encodes all optimal plans as geodesics. In addition, \method is purely discriminative, and focuses only on those features that are relevant towards predicting long-horizon distance relationships.
%
%
%

In the second category are differentiable planning algorithms on a grid world~\cite{Gupta2019,tamar2016value,lee2018gated} and gradient-based planning methods that require supervision through expert demonstration~\citep{upn,dpn}. \Method works in continuous state space, with random and off-policy exploratory data as a pre-training step. Additionally, the metric that \method learns can be used as an intrinsic reward in self-supervised or task-agnostic RL~\citep{warde-farley2018discern,florens2019self,kahn2017self,pong2019self}, to reduce the need of human designed reward. 

Finally, plan2vec builds upon prior methods that plan over a graph with various assumptions~\citep{savinov2018semi,savinov2018curiosity,zhang2018composable,Eysenbach2019SearchOT}. We compare against semi-parametric topological memory~\citep{savinov2018semi}, and show that with a learned value function, plan2vec is able to make more intelligent choices at test time, under limited planning budget.

\section{Experimental Evaluation}\label{sec:results}
\begin{wrapfigure}{r}{0.42\linewidth}
    \centering
    \includegraphics[width=0.30\linewidth]{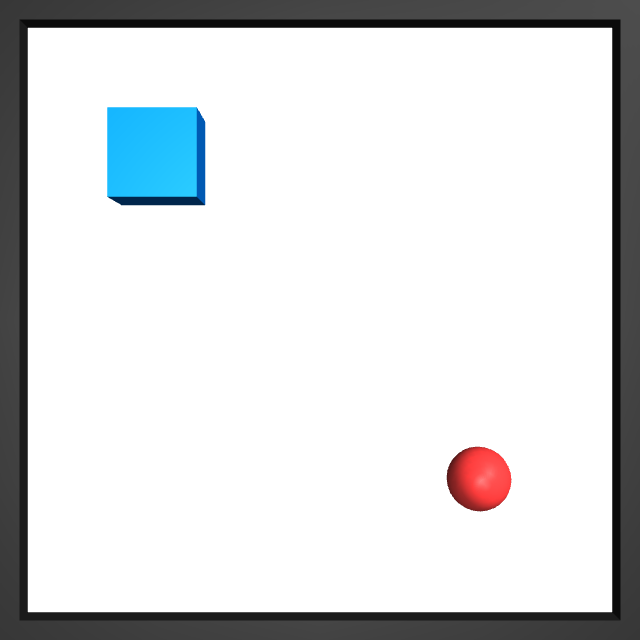}
    \hfill
    \includegraphics[width=0.30\linewidth]{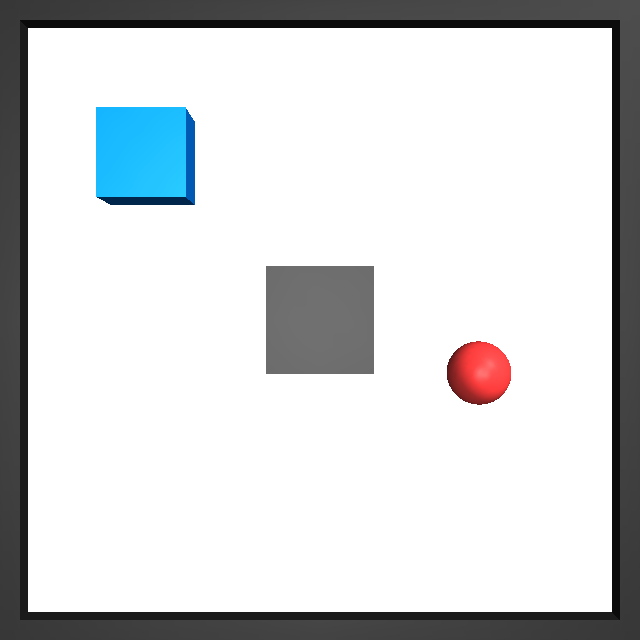}
    \hfill
    \includegraphics[width=0.30\linewidth]{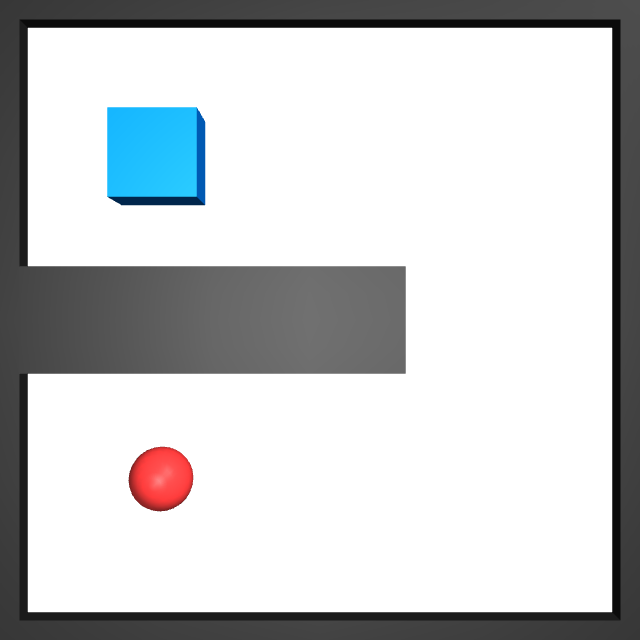}
    \caption{Visual navigation environments: \textit{Open}, \textit{Table}, and \textit{C-Maze}. Agent in {\color{es-blue}blue}. {\color{red}Red sphere} indicates the desired goal.}
    \label{fig:2d_envs}
    \vspace{5pt}
\end{wrapfigure}
In this section, we experimentally answer the following questions:
1) What kind of representation can we learn via planning?
2) How does Dyna-style unroll on the graph affect the sample complexity?
3) Why is graph search needed?
and finally, 4) Would \method work in domains other than navigation, or learn features that are not visually apparent?

To answer these questions, we first examine \method quantitatively on a simulated 2D navigation domain. Then we extend \method to the challenging deformable object manipulation task, were the task is to tie a piece of rope. Finally, we show that \method can learn non-visual features such as the agent's geolocation purely from first-person views without requiring ground-truth GPS data.

\subsection{Simulated Navigation}

\begin{wrapfigure}{r}{0.54\linewidth}
 \begin{minipage}[t]{\linewidth}
    \begin{table}[H]
    \caption{Planning Performance on 2D Navigation}\label{tab:2d_nav_results}
    \vspace{-6pt}
    \centering
     \resizebox{\linewidth}{!}{%
    \begin{tabular}{lccc}
    & \multicolumn{3}{c}{Success Rate (\%)} \\  
            \cmidrule{2-4}
    \multicolumn{1}{c}{\textbf{Image Input}}& Open Room & Table  & C-Maze  \\  
    \midrule
          \method (L2) & \(\mathbf{90.0\pm2.0}\) & \(\mathbf{76.4\pm9.2}\) & \(\mathbf{80.2\pm6.3}\) \\
          SPTM (1-step) & \(39.7\pm6.1\) & \(23.7\pm6.1\) & \(31.4\pm6.5\) \\
                    VAE & \(73.9\pm4.3\) & \(30.2\pm6.5\) & \(52.7\pm5.8\) \\
                 Random & \(3.2\pm2.5\) & \(3.5\pm2.5\) & \(4.7\pm2.8\) \\
        \end{tabular}}
    \end{table}
    \end{minipage}
\end{wrapfigure}
The maze domain is a square, 2-dimensional arena with continuous \((x,y)\) coordinates. A top-down camera view is fed to the robot~(block in {\color{es-blue}blue}). We use ground-truth coordinates for evaluation only. Our experiment covers three room layouts with increasing levels of difficulty: an open room, a room with a table in the middle, and a room with a wall that separates it into two corridors that resembles a C-shaped maze (see~\fig{fig:2d_envs}).

\begin{figure}
\begin{minipage}[t]{0.30\textwidth}
    \centering%
    \raisebox{8pt}[0pt][-5pt]{
        \includegraphics[height=85pt]{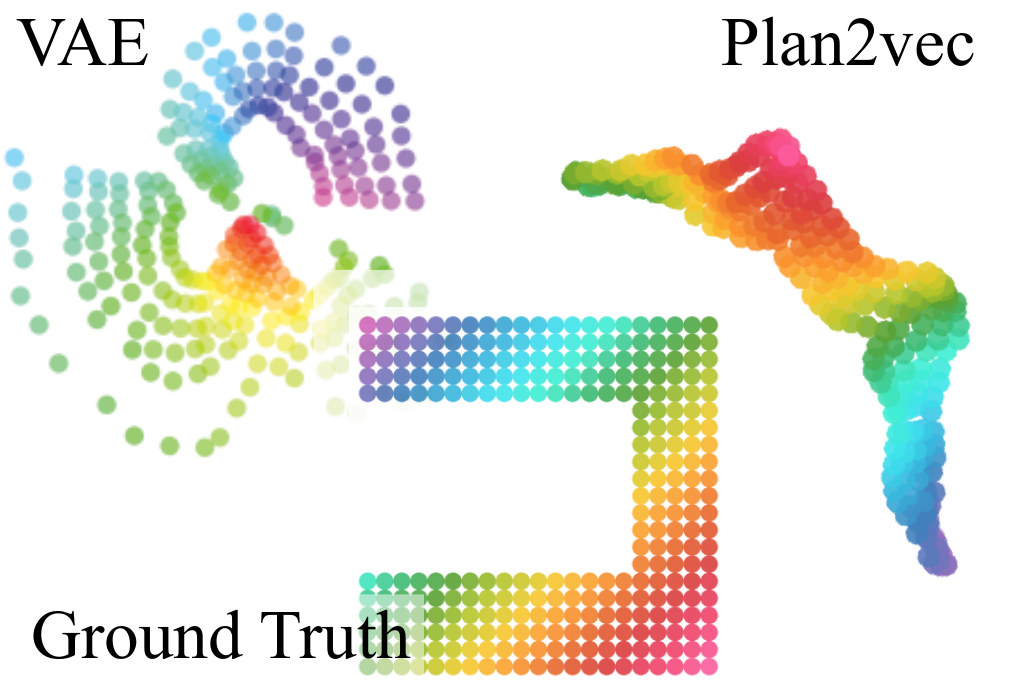}
    }\vspace{-5pt}%
    \caption{\small 2-dimensional latent embedding. \Method's embedding demonstrates clear global structure beyond close neighbors.}
    \label{fig:embedding}
\end{minipage}\hfill%
\begin{minipage}[t]{0.66\textwidth}
    \centering%
    \includegraphics[height=95pt]{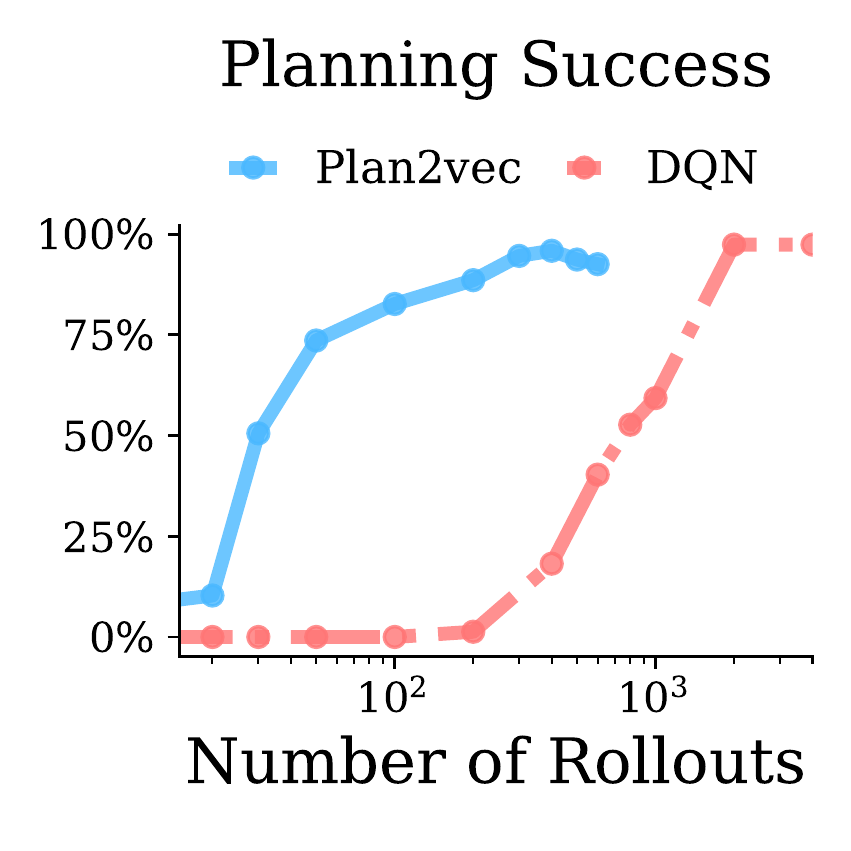}\hfill%
    \includegraphics[height=95pt]{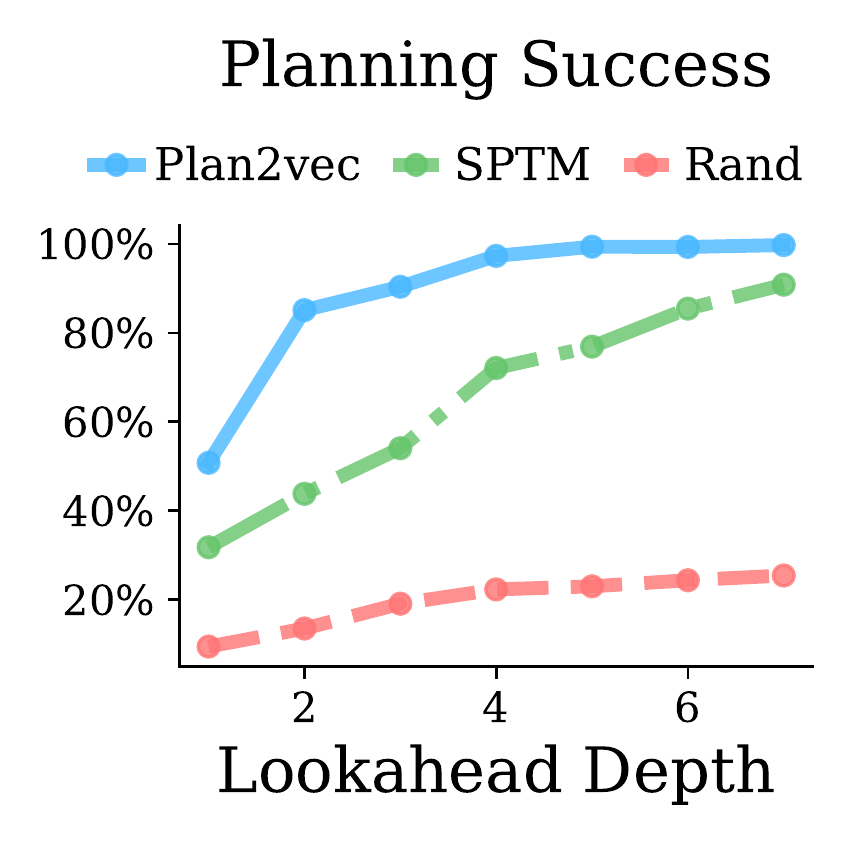}\hfill%
    \includegraphics[height=95pt]{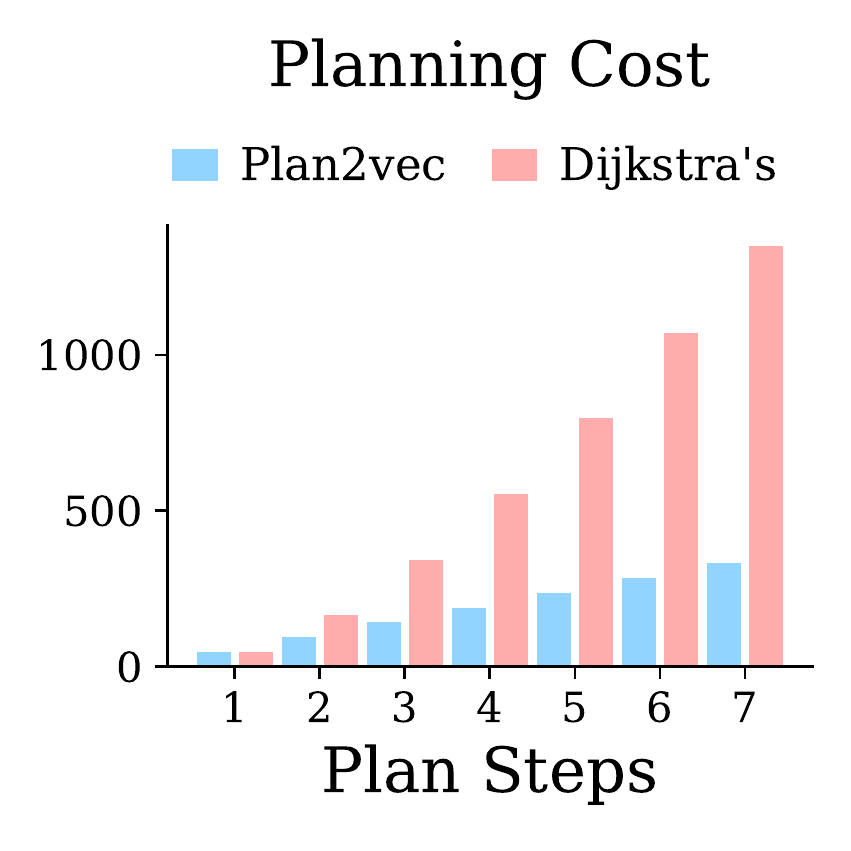}%
    \vspace{-5pt}
    \caption{\small Left: Success rate vs the number of rollouts used for learning, \Method vs DQN. Center: Success rate with k-step planahead, \Method vs SPTM and a random baseline. Right: Planning Cost, Dijkstra's grows quadratically whereas plan2vec is linear. Lower is better.}\label{fig:planning_and_sample_complexity}%
\end{minipage}
\end{figure}

We first qualitatively verify the representation that \method learns by making the latent space 2-dimensional. This allows us to directly visualize the latent vectors by \method against those by a VAE (\citep{kingma2013auto}, see~\fig{fig:embedding}). The embedding VAE learns folds onto itself, whereas \method learns an embedding that respects the overall topological structure of the domain. Furthermore, observations from two opposite ends of the C-Maze are pulled apart, which reflects the longer shortest-path-distance in-between. In other words, \Method embeds optimal plans as roughly straight lines in its learned latent space.

We further study how much data it takes for \method to learn compared to standard off-line reinforcement methods such as fitted Q-iteration~\cite{riedmiller2005neural}. We generate a fixed dataset, then vary the amount given to both \method and a standard deep Q-learning algorithm during training. We plot the planning performance of the learned value function in \fig{fig:planning_and_sample_complexity}a. Both methods achieve \(100\%\) when given sufficient data, but \method requires at least 1 magnitudes less. This encouraging result shows the benefit of learning from a graphical model as opposed to replays from a linear buffer, and \method's ability to efficiently construct optimal plans from off-policy, exploratory experience.

Combination of search and value learning is required in harder domains that requires a strong behavior policy~\citep{silver2017alphaGo,hamrick2019save,anthony2017exit}.  In \fig{fig:planning_and_sample_complexity}b, both \method and SPTM improves in performance with more lookahead search budge, but \method, which distils from a search expert during training, acquires a more informative long-range value estimate and better performance.  When we compare how the cost of finding the shortest path scales with the amount of planning lookahead (see~\fig{fig:planning_and_sample_complexity}c). We found that \method is linear in plan depth, as it amortizes the planning cost from training; whereas Dijkstra's is quadratic.

\subsection{Manipulation of Deformable Objects}\label{sec:results:rope}

\begin{figure}[b]
    \centering
    \includegraphics[width=.9\textwidth]{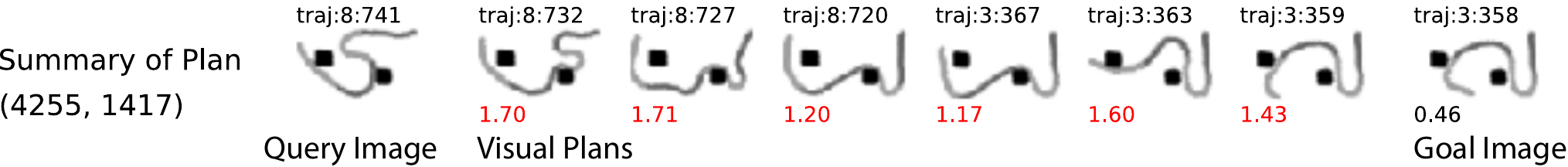}
    \caption{Example of visual plan generated by \method on the Rope Domain showing steps coming from two different trajectories (8 and 3). Each transition only perturbs the configuration of the rope locally. The numbers above denote the trajectory and time step the image is from, the number below represents the score by the local metric \(f_\phi\). Note that the transition from sequence \(8 \rightarrow \) sequence \(3\) occurred in-between the 3rd and 4th step. All the other transitions are real physical transitions.}
    \label{fig:rope_out_traj}
\end{figure}

We now apply \method to learn representations of a deformable object that lacks a structured configuration space. Past methods in this space either rely on learning a generator function~\cite{kurutach2018learning}, or model-free reinforcement learning that can only accomplish a single task~\cite{wu2019learning}. In contrast, \method is purely discriminative, and can generalize to a dynamic set of goals.

We apply our method to a recent rope dataset~\citep{wang2018ropedataset}. This dataset comprises of 18 sequences that include in-total 14k gray scale photos of a piece of rope. Two pegs fixiated on the table impose constraints that need to be respected during each transition.
After training, \method is able to find a visual plan given any pair of start and goal configurations regardless of whether they come from the same trajectory. \fig{fig:rope_out_traj} shows an example of such plans found by \method. Each step only slightly perturbs the configuration of the rope, making the entire plan feasible.

It is difficult to design quantitative evaluation metrics for this domain. For evaluation, we select a start and goal image from the same trajectory, and compare the visual plans made by \method against the ground-truth sequence in-between. We include these additional results in the appendix.

\subsection{Beyond Visual Similarity}

\begin{figure}[tb]
    \centering%
    \newcommand{\hh}{3cm}
    \includegraphics[height=\hh]{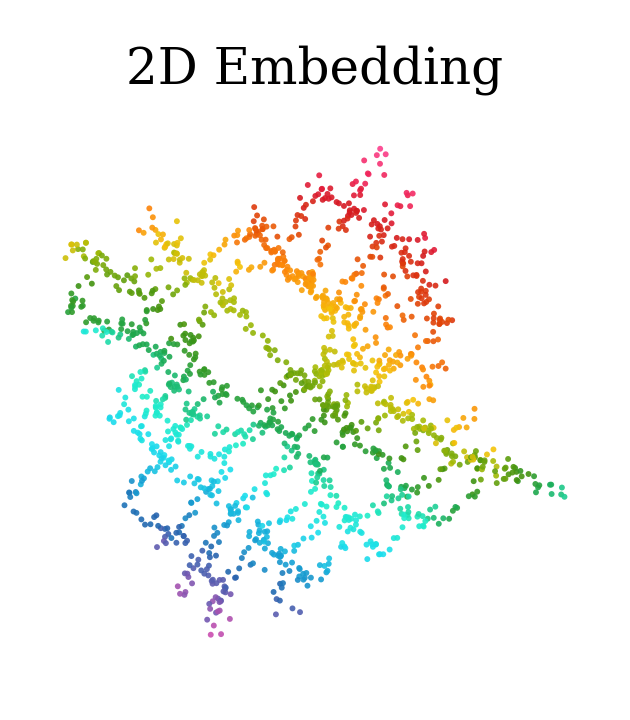}\hfill%
    \includegraphics[height=\hh]{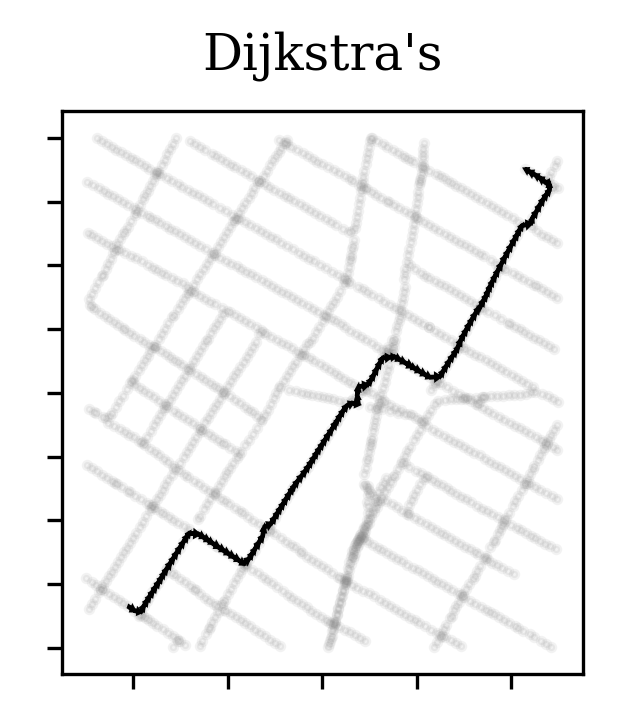}\hfill%
    \includegraphics[height=\hh]{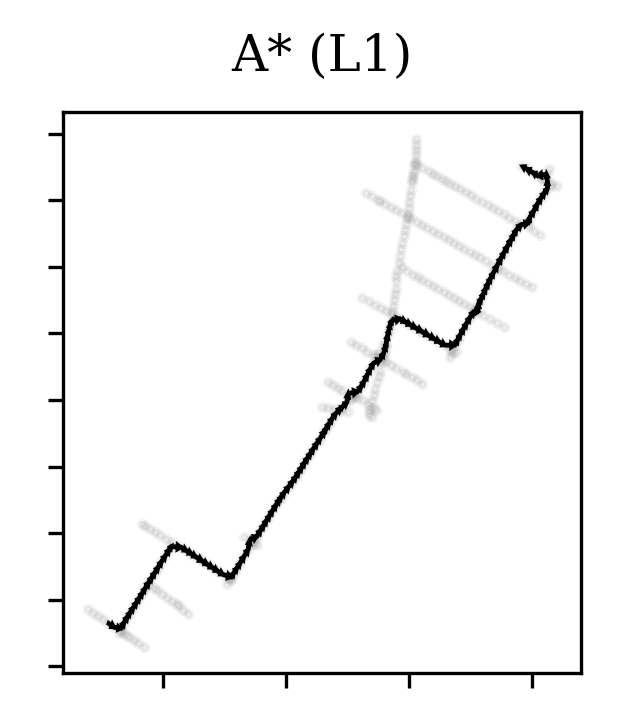}\hfill%
    \includegraphics[height=\hh]{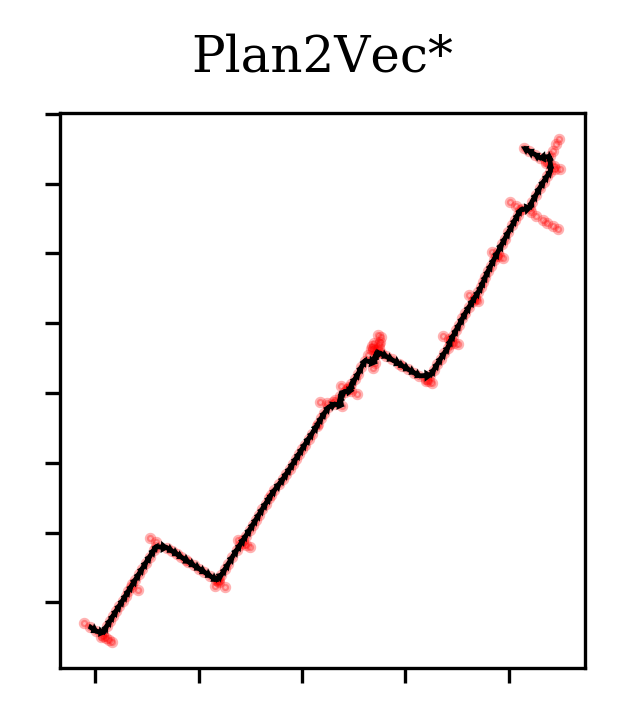}%
    \hspace{1em}%
    \raisebox{-21pt}{%
        \includegraphics[height=3.8cm]{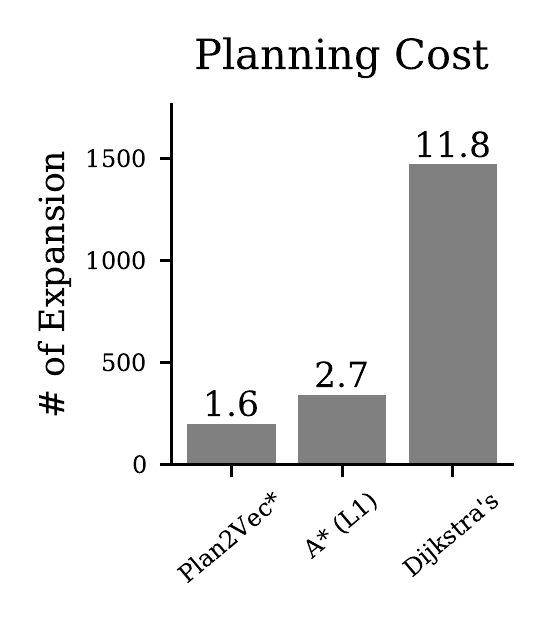}%
    }\vspace{-1em}%
    \caption{Planning Cost. {\color{gray}Gray dots} show the vertices that are expanded during search. We color the expanded nodes with \method in \textcolor{red}{red} to make the few expanded nodes more visible. The plans span \(200\) steps, \(\sim 1.2\) kilometers each. Number on top of bars show the average cost per planning step. With a strong heuristic (\(\ell^1\) distance), A* is more economical than Dijkstra's. But with a learned heuristic \method approaches optimality: a single expansion per step.}\label{fig:streetlearn_planning_cost}
\end{figure}

\begin{wrapfigure}{r}{0.5\textwidth}
    \begin{minipage}[t]{\linewidth}
        \begin{table}[H]
        \caption{
            \small
          \textit{1-step} Planning Performance on StreetLearn. Goals are sampled within 50 steps of the starting point.
        }\label{tab:plan2vec_streetlearn_results}
        \centering
        \vspace{-6pt}
        \resizebox{\linewidth}{!}{%
            \begin{tabular}{lccc}
                            & \multicolumn{3}{c}{Success Rate (\%)} \\  
            \cmidrule{2-4}
        \multicolumn{1}{c}{\textbf{Street Learn}} 
                            & Tiny          & Small      & Medium       \\
            \midrule
            Plan2vec (Ours) & \(\mathbf{92.2\pm2.9}\) 
                            & \(\mathbf{57.2\pm4.3}\) 
                            & \(\mathbf{51.4\pm6.9}\) \\
            SPTM (1-step)   & \(31.5\pm5.8\) 
                            & \(19.3\pm5.8\) 
                            & \(20.2\pm5.2\) \\
            VAE             & \(25.5\pm5.6\) 
                            & \(14.4\pm4.8\) 
                            & \(16.9\pm5.5\) \\
            Random          & \(19.9\pm5.4\)  
                            & \(12.0\pm5.2\)  
                            & \(12.7\pm4.6\) \\
        \end{tabular}}\end{table}%
    \end{minipage}%
\end{wrapfigure}%
In previous domains, visual similarity goes a long way in revealing the distance in the configuration space. Generative models rely on such prior in order to learn, which make them potentially less suitable for learning distance information that are visually inconspicuous. Navigation in a real-world scenario offers a great example -- it is impossible to tell  the direction based off two photos alone. Yet a city resident knows exactly how to navigate from one to another.

We now apply \method to the challenging large scale navigation dataset Street Learn~\citep{streetlearn}. We found that \method's supervised learning objective can learn a high-quality value estimate on a large, \(1.4 k\) subset of Street Learn just under two hours, using only sequences of camera image and step-wise distance without access to the GPS locations. We inspect the learned embedding by restricting the latent space to 2-dimension, and discover a high-quality metric map ~(\fig{fig:streetlearn_planning_cost}a).

Internalizing such a map can speed up planning and improve generalization. In~\fig{fig:streetlearn_planning_cost} we compare the cost of heuristic search with and without using the distance function \method learns. Dijkstra's SPF algorithm expands all nodes in the graph exhaustively, whereas \(A^*\) using the \textit{Manhattan distance} as search heuristic fails to understand the diagonal streets near Broadway. \Method almost optimally captures the shortest-path-distance on this domain, and out performs all other methods.

In \tab{tab:plan2vec_streetlearn_results}, we artificially limit the computation and memory budget for the planner by setting both the lookahead depth \(k\) and the memory size \(\vert \mathcal H\vert\) for the priority queue to 1. In this interesting regime, a good planning heuristic is necessary for good performance. We train an embedding \(\phi(x)\) with each method, then use an \(\ell^2\) metric defined on this embedding as the search heuristic. The VAE baseline barely performs above random. This is expected for unsupervised methods that rely on visual inductive priors for embedding. In comparison to SPTM's 1-step local metric \(d\), \method performs \(2\)-\(3\times\) better consistently across all three datasets.

\subsection{Generalization With A Metric Map}

\begin{figure}[tb]
    \centering
    \newcommand{\hh}{3cm}
    \hspace{-0.4em}%
    \includegraphics[height=\hh]{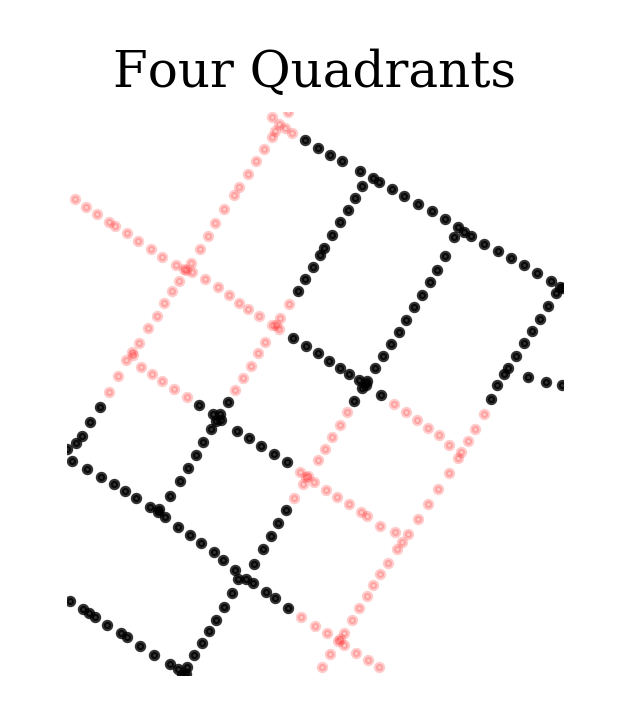}
    \hspace{-0.2em}%
    \includegraphics[height=\hh]{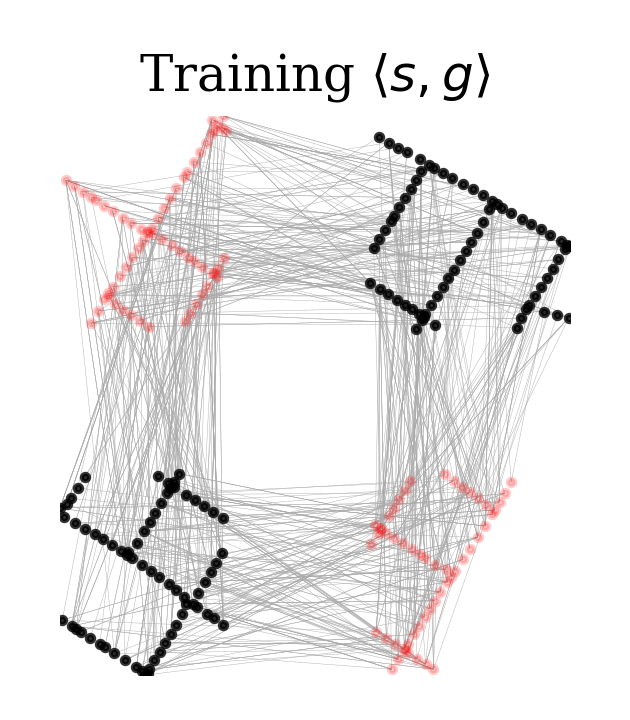}
    \hspace{-0.3em}%
    \includegraphics[height=\hh]{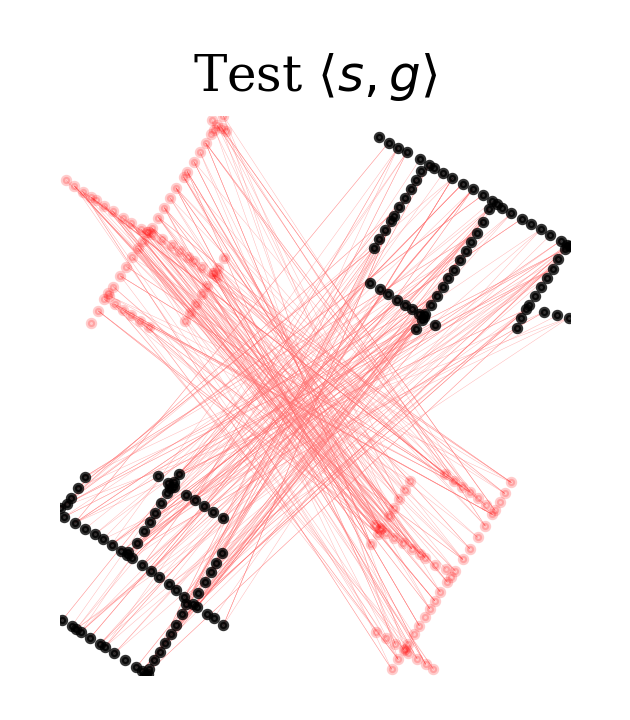}
    \hspace{-0.3em}%
    \includegraphics[height=\hh]{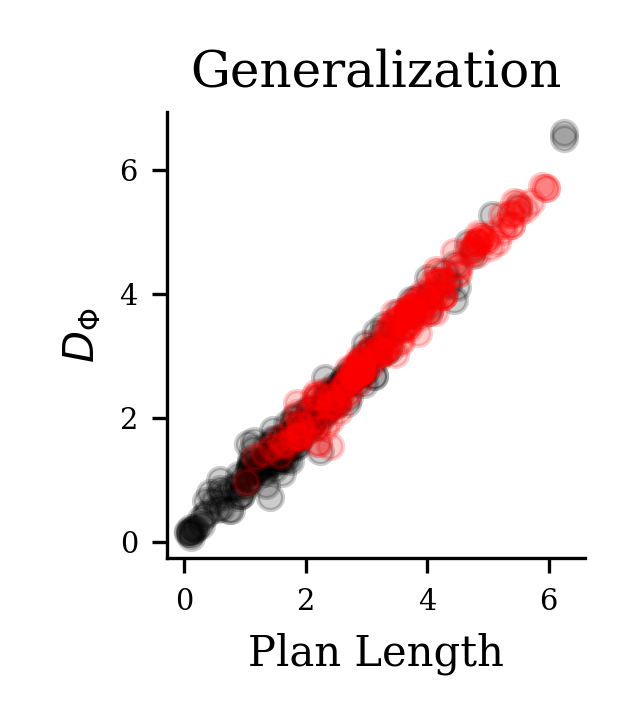}
    \hspace{-0.5em}%
    \includegraphics[height=\hh]{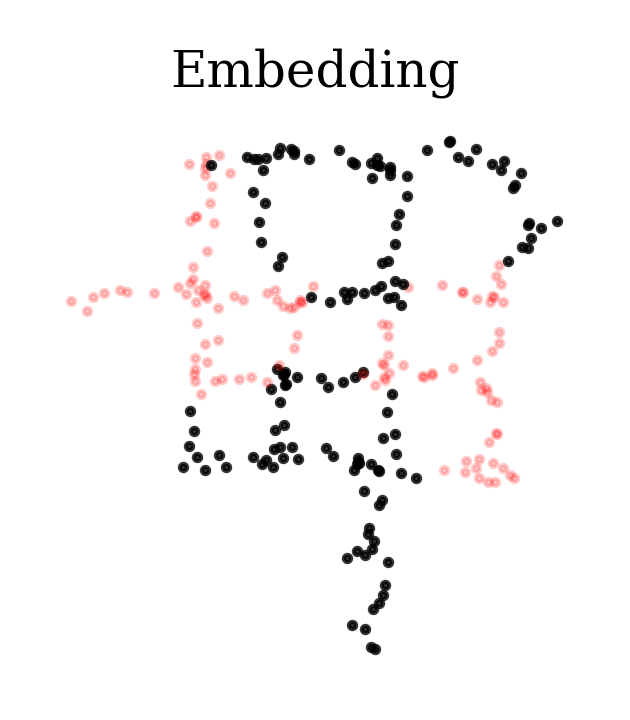}
    \hspace{-0.8em}%
    \includegraphics[height=\hh]{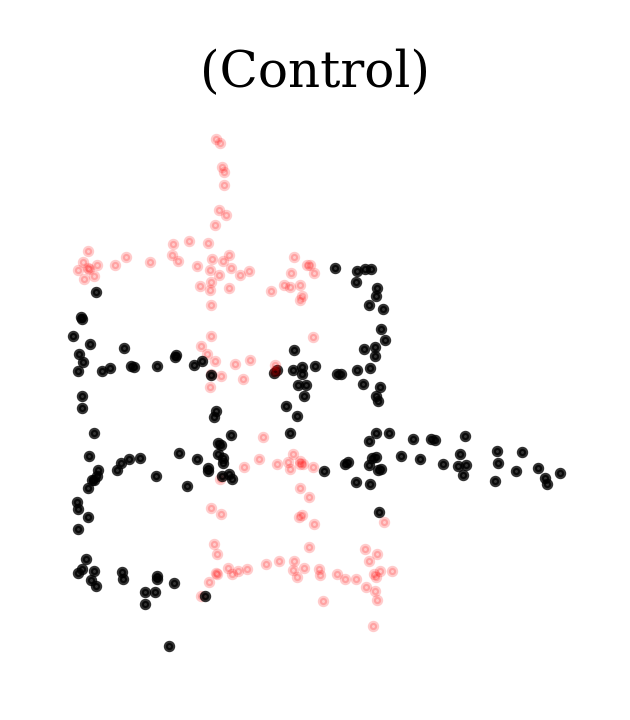}
    \caption{\Method's internal metric map generalize to new tasks. (a) \textbf{\color{pink}Pink} and \textbf{black} color codes the four quadrants (b,\,c) shows the train and test tasks (d) \textbf{\color{red}Red} shows the distance prediction on test set, \textbf{black} on training set. (e,\,f) Embedding learned using this restricted training set is similar to a control using the entire dataset. Orientation of learned embedding depends on seed. Alignment to the axis is due to use of \(p=1.2\). Randomly picked amongst 3 seeds.}
    \label{fig:streetlearn_generalization}
\end{figure}

An important reason to distill plans into a neural network is generalization. In previous experiments, planning generalizes to previously unseen tasks by interpolation. Now we want to ask: how about we remove training tasks that go between large areas of the map -- would \method still able to generalize to bundles of task configurations it has never seen during training? 
In this experiment, we divide the map into four quadrants (\fig{fig:streetlearn_generalization}a) and remove tasks that route between diagonally opposing quadrants during training. To our surprise, the learned embedding is as good as the control that trains with the entire task set. This result depends on the connectivity of the road network, but it shows that \method can sometimes generalize despite of categorical removal of training tasks.


\section{Conclusion}

We have presented a discriminative and unsupervised approach to learn long-horizon distance relationships via planning. Our method does not generate images, is model-agnostic, and requires no access to expert action data. In comparison to model-free reinforcement learning methods that sample directly from the environment, our model-based approach makes more efficient use of otherwise disjoint trajectories. The embedding plan2vec learns encodes the shortest-path between observations as geodesics in the latent space, which reduce iterative planning to fast, parameterized lookup. We demonstrate these desirable properties on one simulated and two challenging real-world datasets, and propose plan2vec as a valuable pre-training step for reinforcement learning agents from off-line exploratory data.


\bibliography{references}

\onecolumn
\setcounter{section}{0}
\renewcommand{\thesection}{\Alph{section}}
\section{Algorithmic Details}\label{sec:algorithmic_details}

We experiment with two variants. The two algorithms differ by the sampling policy and the regression target for the value function. The main variant in \alg{alg:plan2vec_main} uses graph search as the sampling policy and no bootstrapping during learning. Search terminates either when the goal is reached, or it has exhaustively searched the entire graph without finding a path. We precompute the pairwise adjacency matrix so search is instantaneous. To use \(A^*\) in the place of Dijkstra's during training, each node expansion would require pairwise comparison with respect of the goal. \citealt{savinov2018semi} introduced a trick to speed this up by pre-computing all of the distance-to-goal value for a fixed goal. This can be sped up additionally by caching the latent vectors in a Siamese architecture~\cite{savinov2018curiosity}, so that only the kernel operation \(\Vert z_i - z_g\Vert\) needs to be computed.

The second variant, show in \alg{alg:value_iteration} is similar to TreeQN~\cite{farquhar2018treeqn}. It uses breath-first-search (BFS) with a fixed search depth \(k\) at each step before making a greedy selection to minimize \(D(v_{t+1}, v_g)\). When the lookahead depth \(k=1\), this is identical to standard 1-step Q-learning. When \(k>1\) BFS is strictly stronger than \(\epsilon\)-greedy, because it exhaustively finds the neighbor within the \(k\)-step ball around the current vertex as opposed to \(1\)-step neighbors. We use a target network for bootstrapping the values. We found when applied to the graph, this variant is unstable and is sensitive to hyperparameters. Training stability improves with larger \(k\).

\vspace{1em}
\begin{figure}[htb]
\centering
\begin{minipage}{0.85\textwidth}
\input{appendix/alg_1_plan2vec_value_iteration}
\end{minipage}
\end{figure}

\section{Experimental Setup}

We use \(64\times64\) gray-scale images for all domains. 

\subsection{Maze Domain}

We collect data samples in parallel from 20 random policy, for a total of 1000 rollouts (50 each). Each rollout is 10 steps. We train the local metric function for 40 epochs, at a learning rate of \(10^{-4}\) using the Adam optimizer~\cite{kingma2014adam}. To construct training pairs, we set the sample ration to \(1 : 1: 2\) for the label \textit{'identical'}, \textit{'neighbor'}, and \textit{'far-away'}. It is know in the contrastive learning literature that increasing the ratio of negative examples improves learning~\cite{tian2019contrastive}. With ground-truth data, we are able to verify the quality of the local metric function by directly visualizing the off-trajectory neighbors it finds. The accuracy is evaluated over a 10-fold validation set. We latter found that using a negative hinge loss for the third category improves performance, but all results reported here are carried using smoothed \(\ell^1\) loss.

To learn the global metric, during each value iteration, we collect a batch of 20 parallel planning trajectories, 20 steps each. We then run 6 optimization epochs with a batch size of 32 samples per mini-batch. We found this parameter setting perform well.

\subsection{Rope Domain}

Details on the dataset is available in~\cite{wang2018ropedataset}. We consider images separated by \(k=2\) or less as neighbors. This is a hyperparameter that can be adjusted depending on the data. We use a 10-fold train/test split for evaluation. Due to the large size of the pair-wise dataset, we only train for 5 epochs, with a mini-batch size of 16 at learning rate of \(10^{-4}\) using the Adam optimizer~\cite{kingma2014adam}. 

\subsection{StreetLearn}\label{sec:streetlearn_dataset}

We created four subsets from Street Learn that cover increasingly larger areas. We use the smaller three sets for evaluation, and show case the learned metric map with the largest one. The Street Learn dataset is very sparse in that views are around 10 meters apart. For this reason generalization from the local metric is not as critical as for the maze domain, as the majority of the edges come from the sampled transitions.

We calculate the ground distance using the latitude/longitude coordinates multiplied with a Mercator correction factor on the latitude (\(\approx 1.74\)). We scale the 1-step ground distance that is used to construct the graph with a scaling factor. This step is critical because otherwise gradient masking occurs due to the finite precision at those small values. It additionally prevents the mismatch between the initial parameterization of the network and the distribution of the distance targets.

\Method's supervised objective makes learning the complex street topology directly from camera input very computationally efficient. Full convergence on the largest dataset takes just under 2 hours on a single V100 GPU. On the small dataset, the entire training takes 9 minutes.

\vspace{1em}
\begin{figure}[htb]
    \renewcommand\theadgape{\Gape[4pt]}
    \renewcommand\cellgape{\Gape[4pt]}
    \begin{minipage}[t]{\linewidth}
        \begin{table}[H]
        \caption{Details of Street Learn Subsets}\label{tab:streetlearn_dataset}
        \centering
        \vspace{-6pt}
        \begin{tabular}{c|cccc}
            \textbf{Subset} & Tiny            & Small         & Medium        & Large             \\
            \midrule
            Views           & \(53\) 
                            & \(255\) 
                            & \(501\)
                            & \(1495\) \\
            \midrule
            \makecell{
                \vspace{0.5em}
                \textbf{Map Area}\\
                lat, long,\\
                height, width
                \vspace{0.5em}
            }    
                            & \makecell{
                                \(40.72891,\)\\ 
                                \(-73.99694,\)\\ 
                                \(0.00143,\)\\ 
                                \(0.00191\)
                              } 
                            & \makecell{
                                \(40.72731,\)\\ 
                                \(-73.99698,\)\\ 
                                \(0.00349,\)\\ 
                                \(0.00397\)
                                }
                            & \makecell{
                                \(40.72690,\)\\ 
                                \(-73.99798,\)\\ 
                                \(0.00475,\)\\ 
                                \(0.00648\) 
                                }
                            & \makecell{
                                \(40.72601,\)\\ 
                                \(-73.99700,\)\\ 
                                \(0.00799,\)\\ 
                                \(0.01000\)
                                }\\
            \midrule
            Map Area        & \(0.025\,\mathrm{km}^2\) 
                            & \(0.4\, \mathrm{km}^2\) 
                            & \(0.64\,\mathrm{km}^2\)
                            & \(1.6\, \mathrm{km}^2\)\\
        \end{tabular}
        \end{table}%
    \end{minipage}%
\end{figure}
\vspace{1em}

\vspace{1em}
\begin{figure}[htb]
    \renewcommand\theadgape{\Gape[4pt]}
    \renewcommand\cellgape{\Gape[4pt]}
    \begin{minipage}[t]{\linewidth}
        \begin{table}[H]
        \caption{Street Learn Hyper Parameters}\label{tab:streetlearn_hyperparameters}
        \centering
        \vspace{-6pt}
        \begin{tabular}{c|cccc}
            \textbf{Subset} & Tiny       & Small         & Medium        & Large \\
            \midrule
            Scaling         & \(200   \) & \(700 \)   & \(2000\)   & \(4000   \)\\
            Num Epochs      & \(500   \) & \(2k\)   & \(5k\)   & \(10k \)\\
            Batch Size      & \(20    \) & \(100 \)   & \(100 \)   & \(100    \)\\
            Learning Rate   & \(1^{-4}\) & \(3^{-4}\) & \(1^{-5}\) & \(1^{-5} \)\\
          Metric \(\ell^p\) & \(1.2   \) & \(1.2 \)   & \(1.5 \)   & \(2      \)\\
        \end{tabular}
        \end{table}%
    \end{minipage}%
\end{figure}
\vspace{1em}
\section{Additional Results with Maze}
\begin{figure}[H]
    \vspace{1.5em}
    \raisebox{-6pt}[0pt][0pt]{
        \hspace{-5pt}%
        \includegraphics[height=2.9cm]{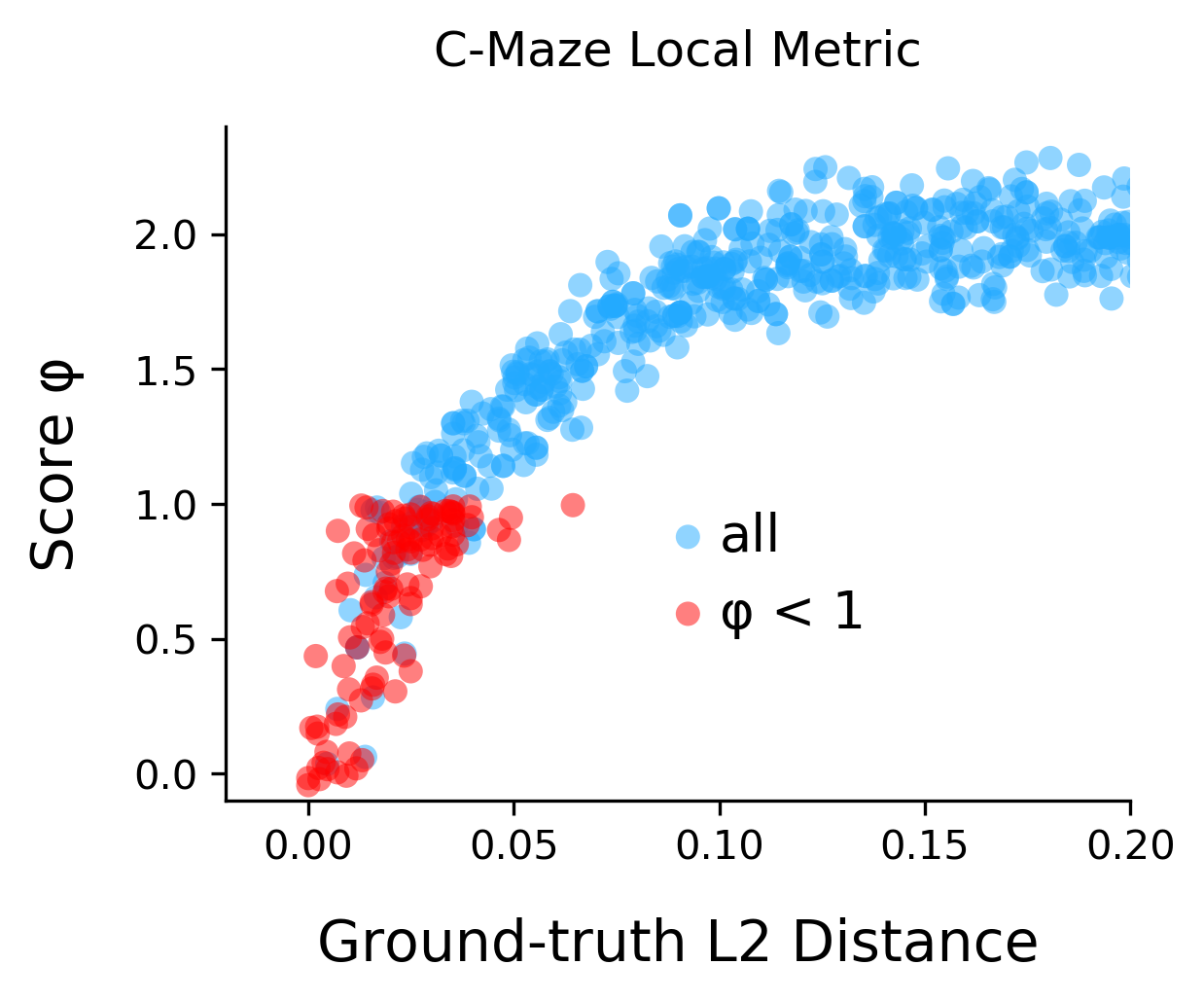}%
        \hspace{-0.8em}
    }
    \raisebox{-6pt}[0pt][0pt]{
        \includegraphics[height=2.9cm]{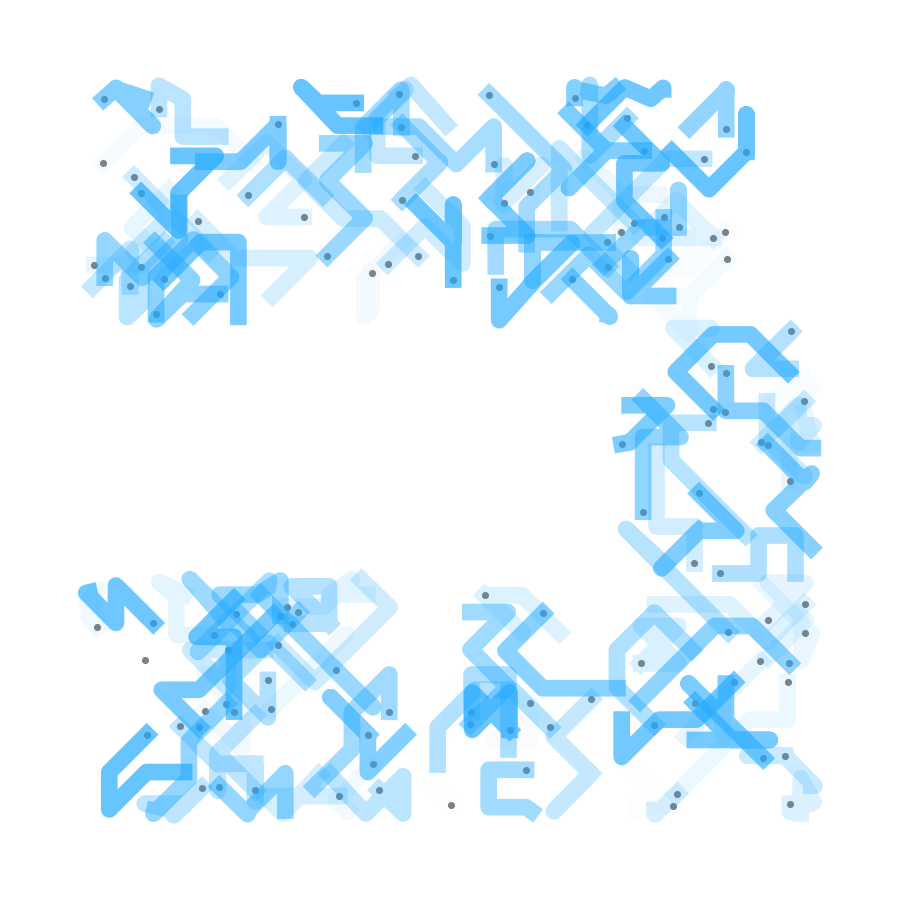}%
        \hspace{-0.8em}
    }
    \raisebox{-6pt}[0pt][0pt]{
        \includegraphics[height=2.9cm]{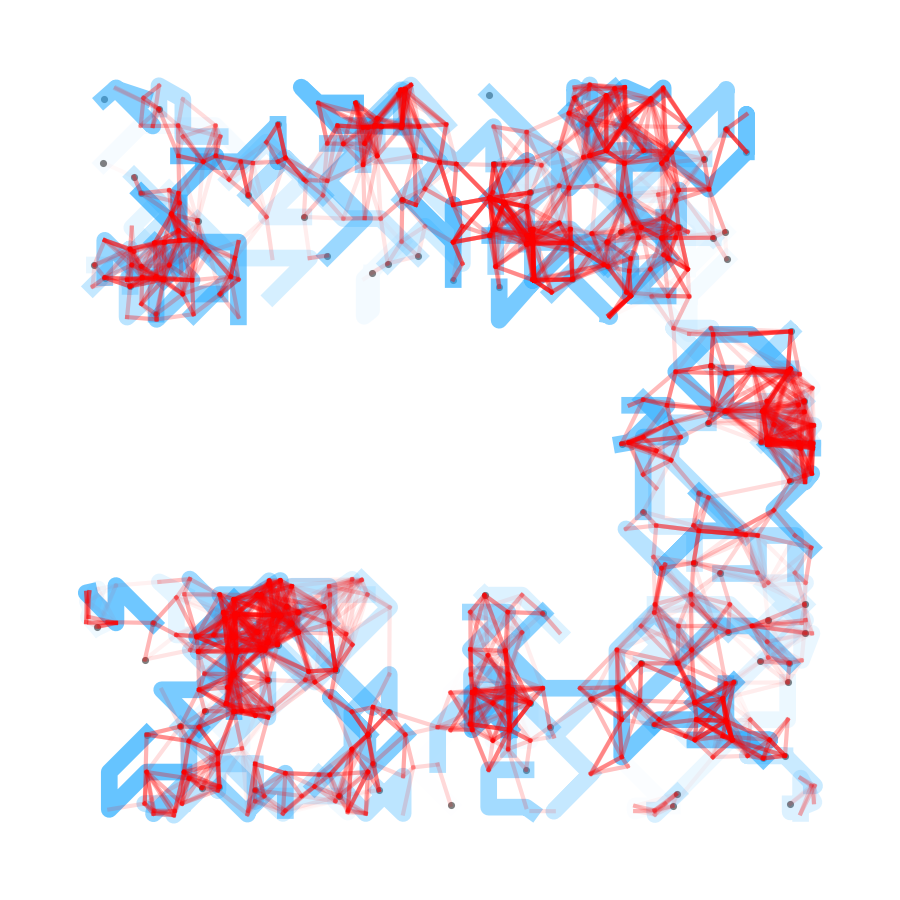}\hspace{0.2em}
    }
    \includegraphics[height=2.6cm]{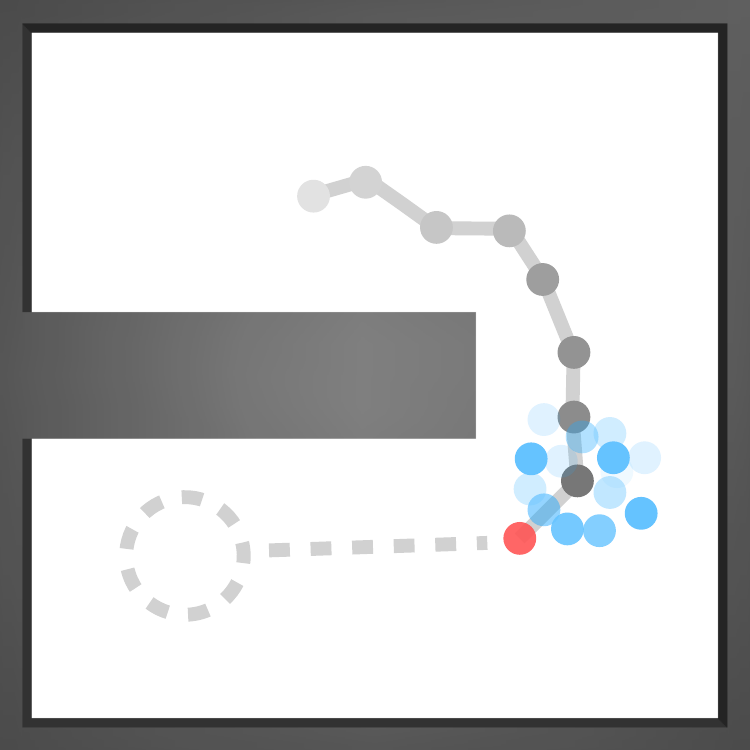}
    \hspace{0.5em}
    \includegraphics[height=2.6cm]{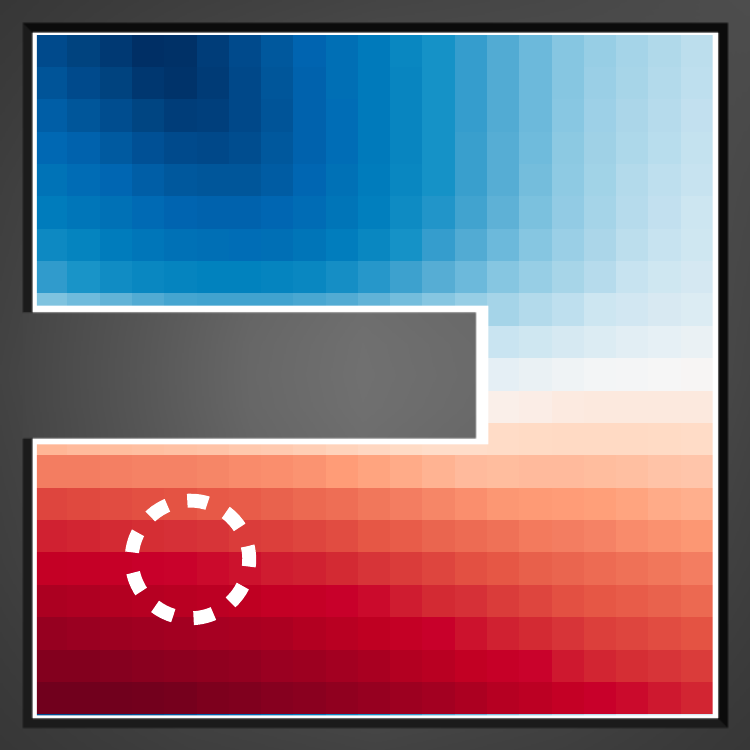}
    \vspace{0.7em}
    \caption{
    (a)~Local metric score in comparison to ground-truth \(L_2\) distance with predicted neighbors in red. We thin the ensemble by \(200\) and \(40\) times to reduce cluttering.
    (b)~Trajectories given in the dataset. 
    (c)~Points from different trajectories are connected by generalizing the local-metric function. Out-of-training-set Connections shown in red.
    (d)~Step sequence in C-Maze, learned via Plan2Vec. Gray dashed circle is the goal position.
        Red dot is the planned next step (1-step), greedy w.r.t the global metric function being learned.
        Blue dots are the neighbors sampled using the local metric function. Gray dot indicates the
        current and past positions of the agent. Sequence shows the agent getting around the wall in
        C-Maze.
    (e)~Learned value function for a goal location on the bottom left corner (white dashed circle). Blue color is further away, red is close. 
    }
    \label{fig:maze_results}
\end{figure}

\fig{fig:maze_results}\,a shows the distribution of the score against ground-truth distance. In shorter ranges, the learned model is able to recover the local metric. But it saturates as the distance increases. We found that aliasing goes down as we increase the dimensionality of the latent space. One can think of this as the network initially emulating a Gaussian random projection, a form of content locality sensitive hashing (LSH, see~\citep{gionis1999hashing}). The score is well-behaved and it is easy to pick suitable values for the neighbor threshold (indicated by the ceiling of the red points). We plot new transitions found by the local metric against those in the dataset~(blue). \fig{fig:maze_results}\,b visualizes the sampled trajectories (in blue, of length 4), whereas \fig{fig:maze_results}\,c shows the new ones found by the learned local metric function.

\section{Additional Results with Rope}\label{sec:rope_additional_results}

\begin{figure}[h]
    \centering
    \begin{minipage}[h]{0.5\textwidth}
        \centering
        \includegraphics[width=0.3\textwidth]{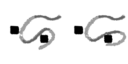}
        \hspace{.1in}
        \includegraphics[width=0.3\textwidth]{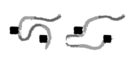}
        \vspace{-5pt}
        \caption{ \small
            Examples of rope pairs that are connected (positive, \textit{left}), 
            and not connected (negative, \textit{right}). 
        }
        \label{fig:rope_pair_example}
    \end{minipage}
    \vspace{5pt}
\end{figure}

\begin{SCfigure}[][h]
  \centering
  \includegraphics[width=0.5\textwidth]{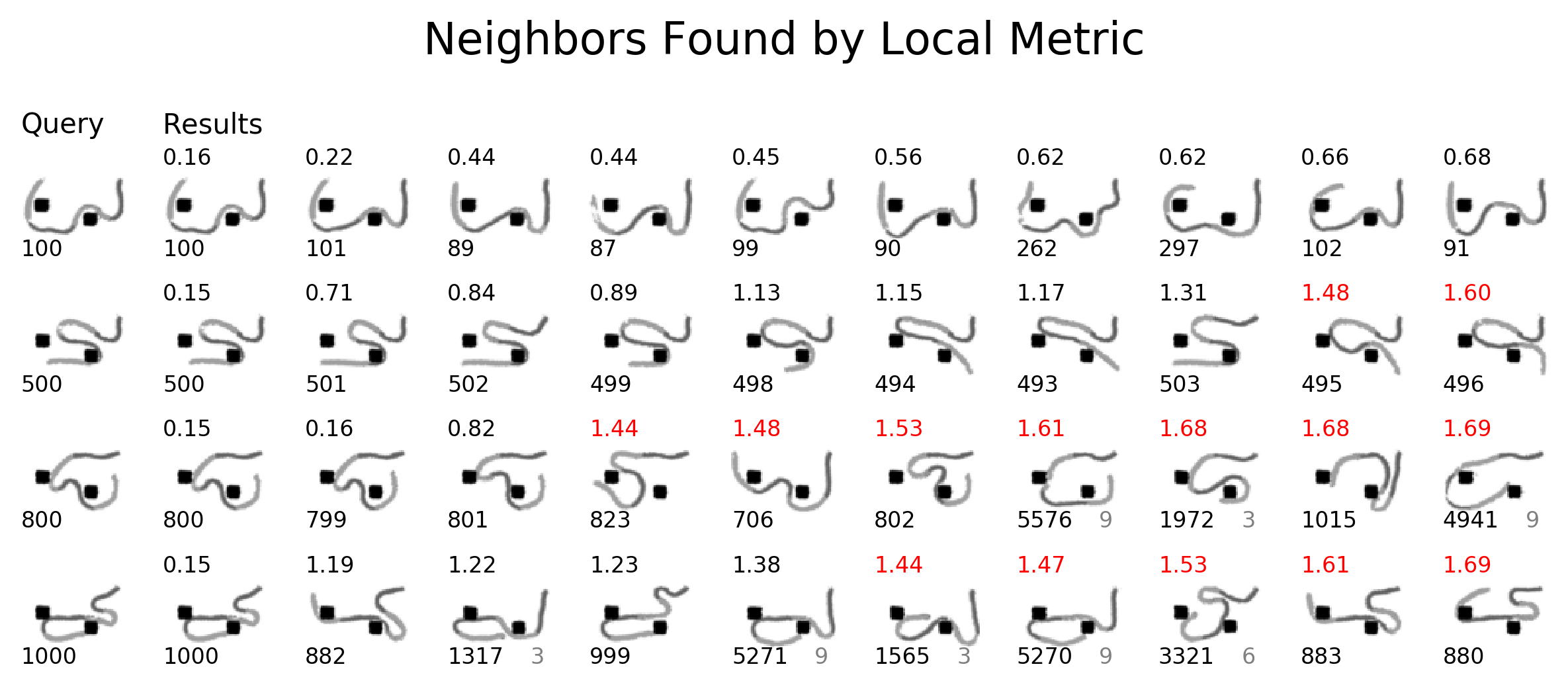}
  \caption{
    \textbf{Neighbors from the rope dataset}. Left-most column are the 
    image used to query for its neighbors in each row. Number on top are
    the local-metric scores; red color indicates the negative examples
    that is above the cut-off threshold of \(1.4\). Number on the bottom 
    shows the index of the image.
}\label{fig:rope_neighbors}
\end{SCfigure}

\begin{figure}[H]
    \vspace{1.5em}
    \includegraphics[trim={.3cm 0.cm 0 .0cm},
    clip,width=.77\linewidth]{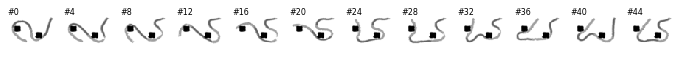}
    \includegraphics[trim={1cm 0.5cm 0 .4cm},
    clip,width=.66\linewidth]{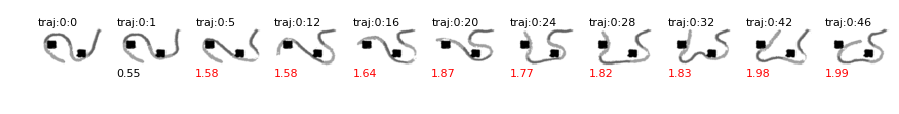}\\
    \includegraphics[trim={1cm 0.5cm 0 .4cm},
    clip,width=1\linewidth]{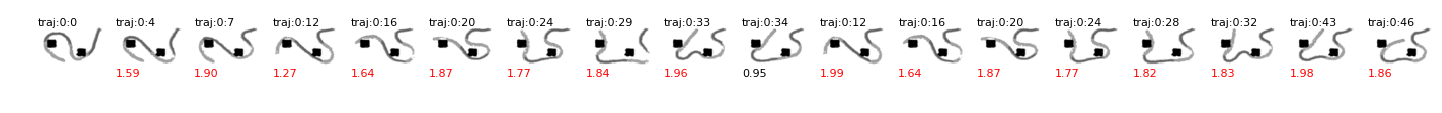} \caption{Examples of trajectories using $o_s$ and $o_g$ randomly sampled from a single trajectory. (\textit{Top}) Original Trajectory, but showing only every fourth frame, (\textit{Middle}) Learned representation + Dijkstra, (\textit{Bottom}) Plan2Vec. Numbers in top left corners denote ground truth trajectory and index of each image, numbers in bottom left are local metric values.
    These planned trajectories are much longer horizon than previously possible with 
    \cite{kurutach2018learning}.
    }
    \label{fig:rope_trajs}
\end{figure}

We show examples of positive and negative pairs for training the local metric in \fig{fig:rope_pair_example}. \fig{fig:rope_neighbors} shows randomly selected images from the dataset versus their top neighbors according to \(d\). \fig{fig:rope_trajs} shows a particular trajectory from the rope dataset, versus a plan found by Dijkstra's shortest-path search algorithm using a local metric \(d\), and one found by plan2vec after training.

\noindent

\section{Additional Results on Street Learn}

\begin{figure}[b]
    \centering
    \newcommand{\hh}{3cm}
    \hspace{-0.4em}%
    \includegraphics[height=\hh]{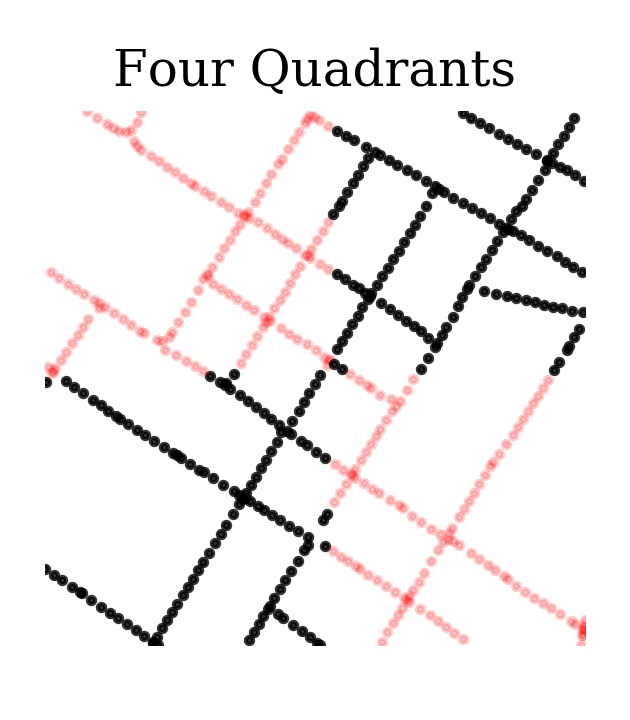}
    \hspace{-0.2em}%
    \includegraphics[height=\hh]{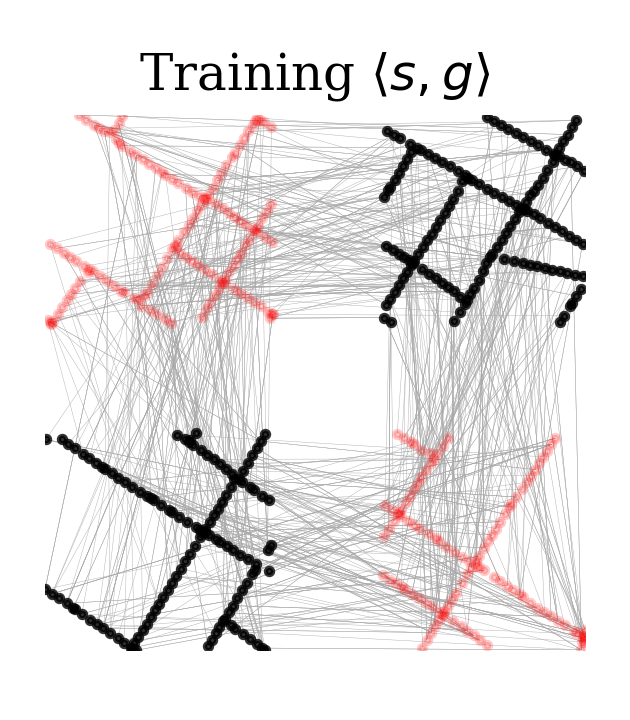}
    \hspace{-0.3em}%
    \includegraphics[height=\hh]{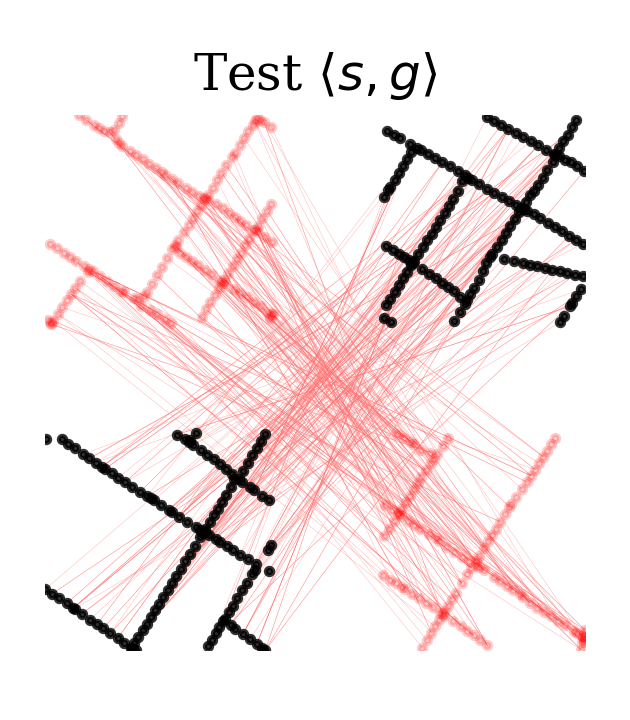}
    \hspace{-0.3em}%
    \includegraphics[height=\hh]{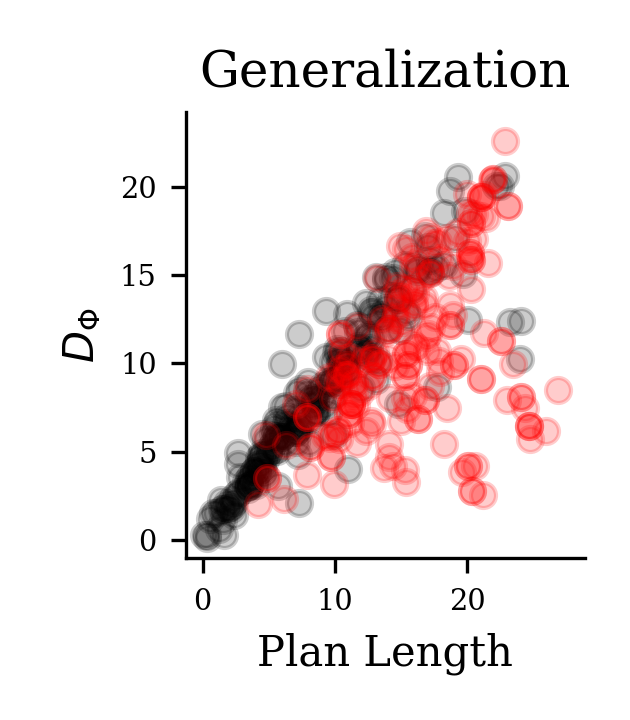}
    \hspace{-0.5em}%
    \includegraphics[height=\hh]{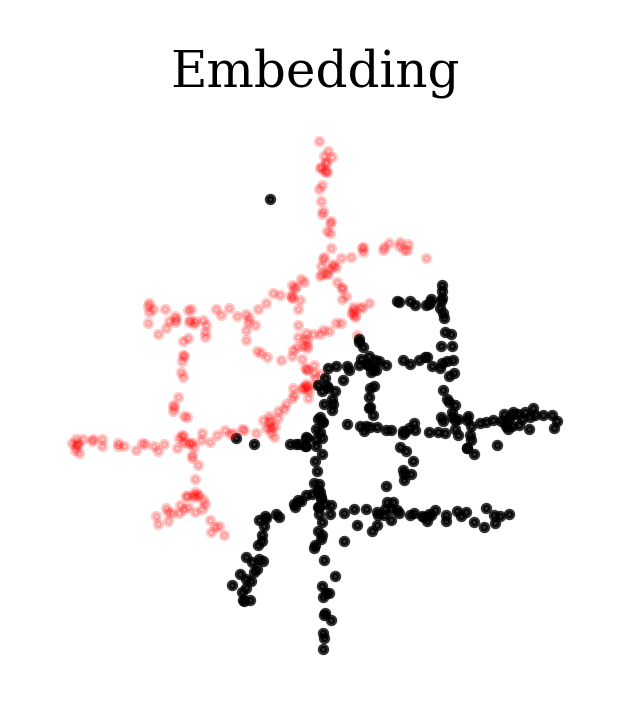}
    \hspace{-0.8em}%
    \includegraphics[height=\hh]{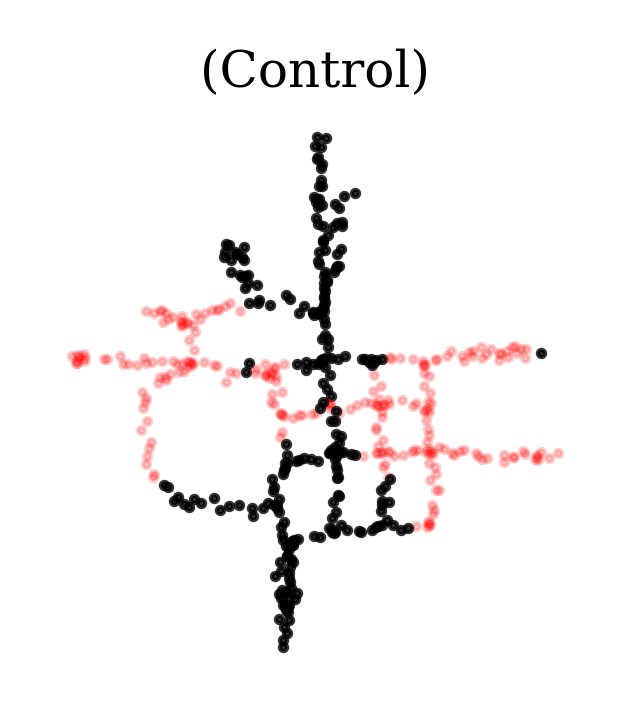}
    \caption{When large bundles of tasks are missing during training, \method underestimates the distance in-between ({\color{pink}Pink} task pairs and {\color{red}red} markers in (c) and (d)). This result shows the importance in using long-horizon plans as learning targets when learning distances. Control in (f) learns from all task-configurations.}
    \label{fig:streetlearn_generalization_appendix}
\end{figure}

Bellow we show additional results on generalization. In a larger map area, removal of diagonal bundles of task configurations during training results in under estimation of the distances in between (see~\fig{fig:streetlearn_generalization_appendix}d). This ablation study shows that to learn metric information between observations that are far apart, long-horizon plans between those areas have to be involved during training. In Q-learning, this is accomplished by iterative value-bootstrapping, which is a much slower.

\subsection{3-dimensional Latent Space}

We include additional visualization of the street map on a 3-dimensional latent space. 

\vspace{1em}
\begin{figure}[h]
    \centering
    \newcommand{\hh}{3cm}
    \includegraphics[height=\hh]{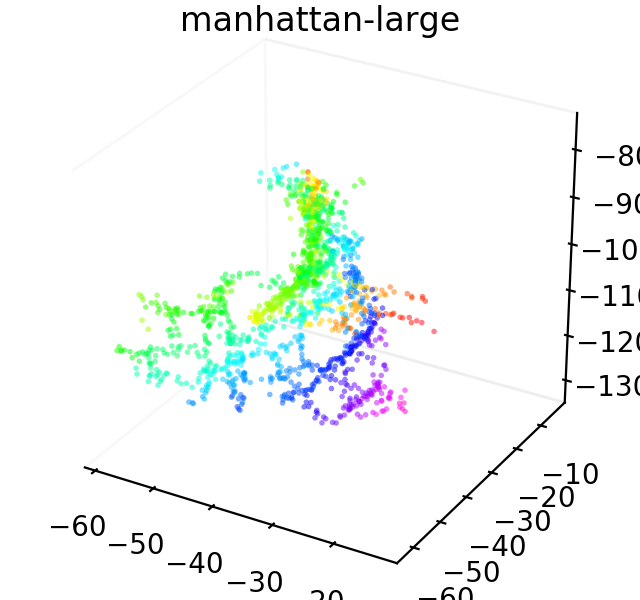}
    \includegraphics[height=\hh]{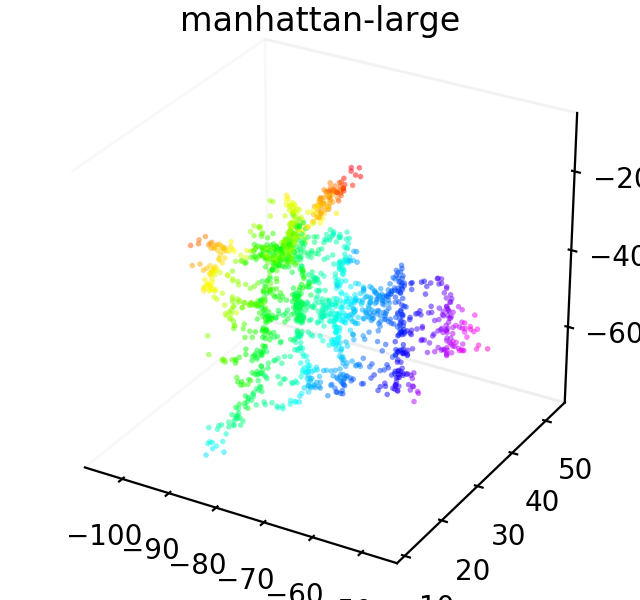}
    \includegraphics[height=\hh]{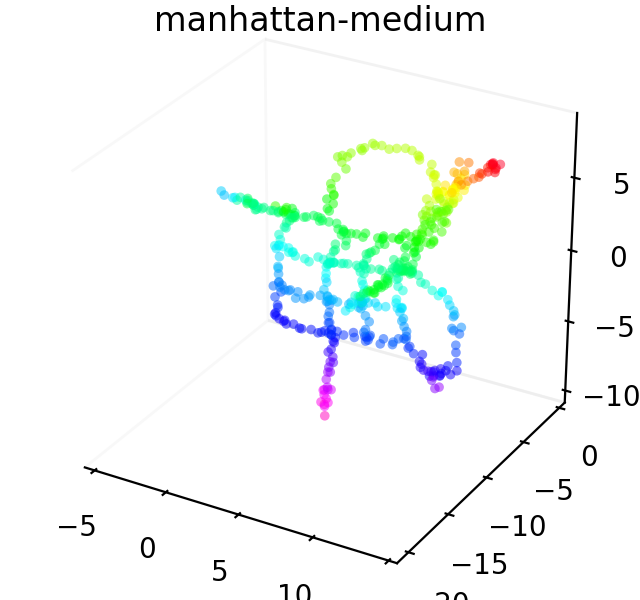}
    \includegraphics[height=\hh]{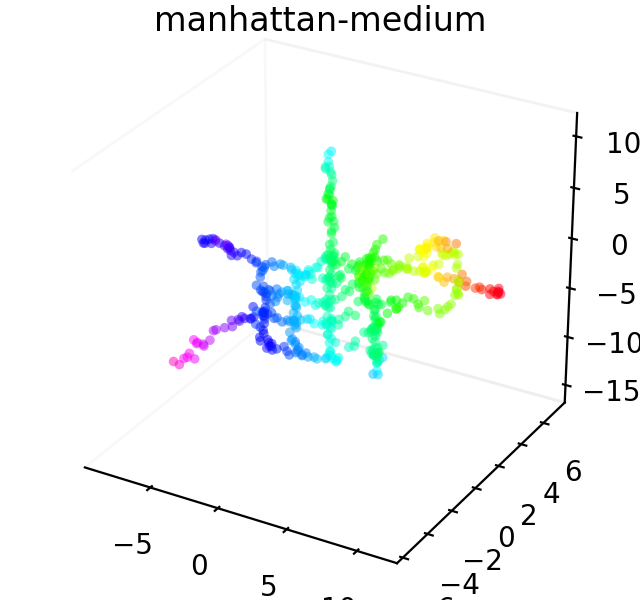}
    \includegraphics[height=\hh]{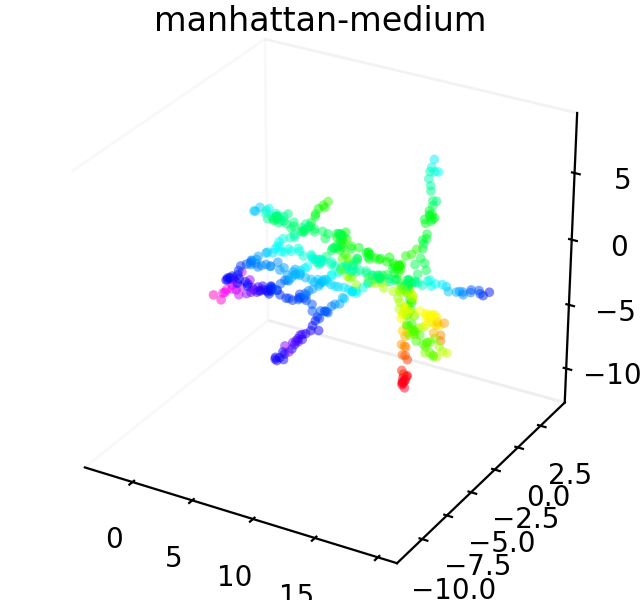}
    \includegraphics[height=\hh]{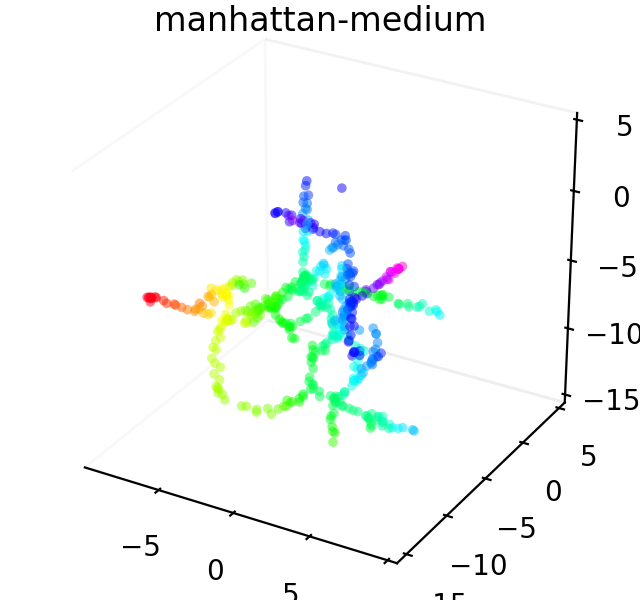}
    \includegraphics[height=\hh]{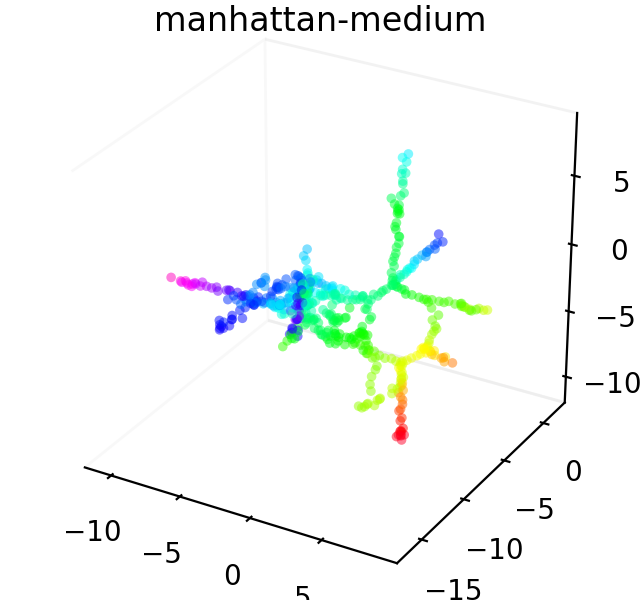}
    \caption{Additional visualization of learned embeddings in a 3-dimensional latent space. (a,b) Manhattan-large; (c-g) Manhattan-medium. \(p=1.2\), showing results from all random seeds. Not cherry-picked.}
\end{figure}

\section{Additional Results on Local Metric \boldmath$d$}

In this section we list the training and test accuracy of the local metric for all domains. \Method is model-based, and is sensitive to ``worm-hole'' connections that are misidentified by the local metric. For this reason, improvements to the local metric and the graph will translate into improvements of the metric \(D\).

\noindent
\begin{minipage}[b]{\textwidth}
    \centering
    \vspace{1em}
    \begin{tabular}{rcc}
                 & \multicolumn{2}{c}{Accuracy \((\pm 25 \%)\)}  \\ 
    Domain       & Train                 & Test                  \\ 
    \midrule
    Open         & \(98.6\% \pm 0.7 \)   & \(97.8 \%  \pm 0.3\)  \\ 
    Table        & \(98.3\% \pm 1.1 \)   & \(97.6 \%  \pm 0.3\)  \\
    Wall         & \(98.3\% \pm 1.2 \)   & \(97.4 \%  \pm 0.4\)  \\
    Rope         & \(96.6\% \pm 0.9 \)   & \(92.4 \%  \pm 1.5\)  \\
    Street Learn & \(99.2\% \pm 0.5 \)   &          -            \\
    \end{tabular}
    \captionof{table}{
        Prediction accuracy for 1-step neighbors on all domains when threshold is set to \(1.5\), right in-between 1 and 2. The local metric function can detect neighbors consistently. Results are averaged over 5 seeds. On Street Learn we use all samples from \textit{Manhattan-large} during training due to the sparsity of the view points in comparison to the large map size.
    }\label{tab:local_metric_accuracy}
\end{minipage}

\section{Architectural Details}

\Method is model-agnostic and can work with a variety of different architectures. We list the details of the network used during our experiments below in the form of pseudocode.

\paragraph{Maze Local Metric} is a five-layer convolution network. We stack the two input images channel wise.
\begin{minted}{python}
LocalMetricConvLarge(
  (trunk): Sequential(
    (0): Conv2d(2, 32, kernel_size=(4, 4), stride=(2, 2))
    (1): BatchNorm2d(32, eps=1e-05, momentum=0.1, affine=True)
    (2): ReLU()
    (3): Conv2d(32, 64, kernel_size=(4, 4), stride=(2, 2))
    (4): BatchNorm2d(64, eps=1e-05, momentum=0.1, affine=True)
    (5): ReLU()
    (6): Conv2d(64, 64, kernel_size=(4, 4), stride=(2, 2))
    (7): BatchNorm2d(64, eps=1e-05, momentum=0.1, affine=True)
    (8): ReLU()
    (9): Conv2d(64, 32, kernel_size=(4, 4), stride=(2, 2))
    (10): BatchNorm2d(32, eps=1e-05, momentum=0.1, affine=True)
    (11): ReLU()
    (12): View(-1, 128)
    (13): Linear(in_features=128, out_features=128, bias=True)
    (14): ReLU()
    (15): Linear(in_features=128, out_features=100, bias=True)
    (16): ReLU()
    (17): Linear(in_features=100, out_features=1, bias=True)
  )
)
\end{minted}

\paragraph{Maze Global Metric} We increase the capacity of the network for the global metric, and adopt a Siamese architecture with an \(\ell^2\)-metric head.

\begin{minted}{python}
GlobalMetricConvL2(
  (embed): Sequential(
    (0): Conv2d(1, 128, kernel_size=(7, 7), stride=(1, 1))
    (1): BatchNorm2d(128, eps=1e-05, momentum=0.1, affine=True)
    (2): ReLU()
    (3): Conv2d(128, 256, kernel_size=(7, 7), stride=(1, 1))
    (4): BatchNorm2d(256, eps=1e-05, momentum=0.1, affine=True)
    (5): ReLU()
    (6): Conv2d(256, 256, kernel_size=(7, 7), stride=(2, 2))
    (7): BatchNorm2d(256, eps=1e-05, momentum=0.1, affine=True)
    (8): ReLU()
    (9): Conv2d(256, 256, kernel_size=(7, 7), stride=(2, 2))
    (10): BatchNorm2d(256, eps=1e-05, momentum=0.1, affine=True)
    (11): ReLU()
    (12): Conv2d(256, 256, kernel_size=(7, 7), stride=(2, 2))
    (13): BatchNorm2d(256, eps=1e-05, momentum=0.1, affine=True)
    (14): ReLU()
    (15): Conv2d(256, 256, kernel_size=(2, 2), stride=(1, 1))
    (16): ReLU()
    (17): View(-1, *(256,))
    (18): Linear(in_features=256, out_features=2, bias=True)
  )
  (head): Lambda(a, b) => norm(a - b, p=2)
)
\end{minted}

\paragraph{Rope and Street Learn} uses the same local metric function. We stack two gray-scale images together into a 2-channel image for the local metric.

\begin{minted}{python}
LocalMetric(
  (trunk): Sequential(
    (0): Conv2d(2, 128, kernel_size=(4, 4), stride=(2, 2))
    (1): BatchNorm2d(128, eps=1e-05, momentum=0.1, affine=True)
    (2): ReLU()
    (3): Conv2d(128, 128, kernel_size=(4, 4), stride=(2, 2))
    (4): BatchNorm2d(128, eps=1e-05, momentum=0.1, affine=True)
    (5): ReLU()
    (6): Conv2d(128, 128, kernel_size=(4, 4), stride=(1, 1))
    (7): BatchNorm2d(128, eps=1e-05, momentum=0.1, affine=True)
    (8): ReLU()
    (9): Conv2d(128, 128, kernel_size=(4, 4), stride=(1, 1))
    (10): BatchNorm2d(128, eps=1e-05, momentum=0.1, affine=True)
    (11): ReLU()
    (12): Conv2d(128, 128, kernel_size=(4, 4), stride=(1, 1))
    (13): BatchNorm2d(128, eps=1e-05, momentum=0.1, affine=True)
    (14): ReLU()
    (15): Conv2d(128, 128, kernel_size=(4, 4), stride=(1, 1))
    (16): BatchNorm2d(128, eps=1e-05, momentum=0.1, affine=True)
    (17): ReLU()
    (18): View(-1, *(512,))
    (19): Linear(in_features=512, out_features=128, bias=True)
    (20): ReLU()
    (21): Linear(in_features=128, out_features=100, bias=True)
    (22): ReLU()
    (23): Linear(in_features=100, out_features=1, bias=True)
  )
)
\end{minted}
To learn the global metric \(D\), we use a ResNet18 trunk \cite{He_2016_CVPR} with an \(\ell^p\) metric head. The network is instantiated with the following pseudo code:
\begin{minted}{python}
# We use the ResNet18 from torchvision.
ResNet18L2(
  (embed): Sequential(
    (resnet_18): ResNet18([2, 2, 2, 2])
    (conv_1): Conv2d(1, 64, kernel_size=7, stride=2, padding=3, bias=False)
  )
  (head): Lambda(a, b) => norm(a - b, p)
)
\end{minted}
\end{document}